%% file: _-acl2023.tex
\pdfoutput=1

\documentclass[11pt]{article}

\usepackage{ACL2023}

\usepackage{times}
\usepackage{latexsym}

\usepackage{bbding}
\usepackage{amsmath}
\usepackage{amssymb}
\usepackage{arydshln}
\usepackage{bbm}
\usepackage{booktabs}
\usepackage{CJKutf8}
\usepackage{combelow}
\usepackage{enumitem}
\usepackage{epsfig}
\usepackage{fancybox}
\usepackage{float}
\usepackage{graphicx}
\usepackage{hyperref}
\usepackage{multirow}
\usepackage{multicol}
\usepackage{pgfplots}
\usepackage{pifont}
\usepackage{setspace}
\usepackage{tabularx}
\usepackage{url}
\usepackage{xspace}
\usepackage{caption}
\usepackage{subcaption}

\usepackage[T1]{fontenc}

\usepackage[utf8]{inputenc}

\usepackage{microtype}

\usepackage{inconsolata}

\usepackage{multirow}

\newcommand{\ssymbol}[1]{^{\@fnsymbol{#1}}}

\newcommand{\model}{\textsc{InBedder}\xspace}

\title{Answer is All You Need: Instruction-following Text Embedding via Answering the Question}

\author{Letian Peng$^1$\thanks{$\ $  The first two authors contributed equally to this work.}, Yuwei Zhang$^1$\footnotemark[1], Zilong Wang$^1$ \\ \bf Jayanth Srinivasa$^2$, Gaowen Liu$^2$, Zihan Wang$^1$\thanks{$\ $  Corresponding authors.}, Jingbo Shang$^1$\footnotemark[2] \\
University of California, San Diego$^1$ \quad Cisco$^2$ \\
  \texttt{\{lepeng, yuz163, zlwang, ziw224, jshang\}@ucsd.edu}\\
  \texttt{\{jasriniv, gaoliu\}@cisco.com}
  }

\begin{document}
\maketitle
\begin{abstract}
\input{0-abstract}
\end{abstract}

\section{Introduction}

\input{1-introduction}

\section{Related Works}

\input{2-related_work}

\section{Problem Formulation}

\input{3-preliminary-yuwei}\label{sec:preliminary}

\section{Methodology}

\input{4-method-yuwei}

\section{Experiments}\label{sec:experiment}

\input{5-experiment}

\section{Conclusions and Future Work}

\input{6-conclusion}

\section*{Limitations}

\input{7-limitations}

\bibliography{custom}
\bibliographystyle{acl_natbib}

\clearpage

\appendix
\input{8-appendix}

\end{document}

%% file: 0-abstract.tex
This work aims to build a text embedder that can capture characteristics of texts specified by user instructions.
Despite its tremendous potential to deploy user-oriented embeddings, none of previous approaches provides a concrete solution for it.
This paper offers a new viewpoint, which treats the instruction as a \emph{question} about the input text and encodes the \emph{expected answers} to obtain the representation accordingly. 
Intuitively, texts with the same (implicit) semantics would share similar answers following the instruction, thus leading to more similar embeddings.
Specifically, we propose \model that instantiates this embed-via-answering idea by only fine-tuning language models on abstractive question answering tasks.
\model demonstrates significantly improved instruction-following capabilities according to our proposed instruction awareness tests and instruction robustness tests, when applied to both large language models (LLMs) (e.g., \texttt{llama-2-7b}) and smaller encoder-based LMs (e.g., \texttt{roberta-large}).
Additionally, our qualitative analysis of clustering outcomes, achieved by applying different instructions to the same corpus, demonstrates a high degree of interpretability.


%% file: 1-introduction.tex
\begin{figure*}
    \centering
    \includegraphics[width=\linewidth]{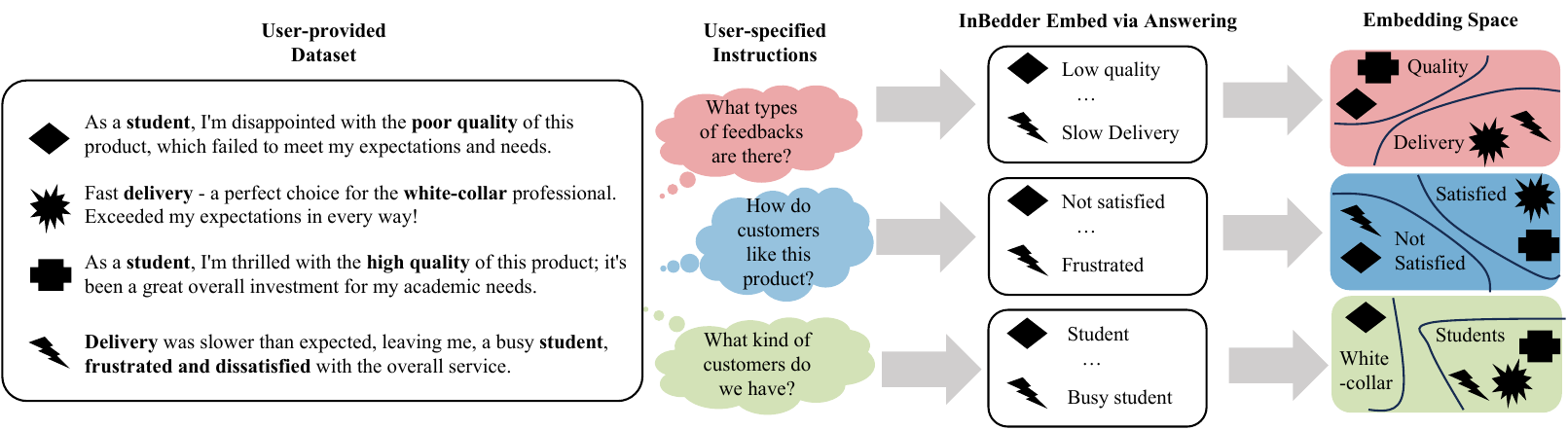}
    \caption{An example workflow of \model. \model takes in both user-provided dataset and user-specified instructions to produce personalized clusterings from which the user can extract insights about the dataset.}
    \label{fig:intro}
    \vspace{-5mm}
\end{figure*}

Text embedders play a crucial role in large-scale textual data analysis and management.
While existing models~\cite{reimers-2019-sentence-bert,SimCSE,ni-etal-2022-sentence,GTR,E5,BGE} demonstrate strong effectiveness in representing texts in general, they lack the ability to address user-specific objectives.
This limitation hinders their application in more intricate scenarios where the embedding task requires the model to represent particular characteristics of the texts~\cite{goalex,zhang-etal-2023-clusterllm}.
Consider Figure~\ref{fig:intro}, where a single set of reviews is required to be clustered in three distinct manners to derive meaningful insights. In response, we attempt to equip the text embedders with instruction-following capability in this paper.

One straightforward solution is to embed the concatenated instruction and input.
Nonetheless, generic textual embeddings represent the texts in a form that can be used in textual similarity tasks, search and clustering, etc, rather than following instructions.
Even for those that are trained with multi-task contrastive objective~\cite{instructor}, there are no guarantee to generalize to new instructions due to the inevitably restricted diversity of training instructions written by humans.

We offer a novel viewpoint, which treats the instruction as a \textbf{question} about the input text and encodes the \textbf{expected answers}.
Specifically, using the instructed input as the \textbf{prompt} to generative language models, we argue that the generated answers can be natively utilized to model semantic similarity under different instructions. 
For instance, given the sentences ``\emph{I love cats}'' and ``\emph{I love dogs}'',
the instruction ``\emph{Do they love animals?}'' will lead to a uniform response of ``\emph{Yes}/\emph{Certainly}/...''; Conversely, distinct answers would be generated in response to ``\emph{What animals do they love?}''.
Therefore, we believe that the expectation of answer representations given the prompt can serve as an instruction-following embedding.
We support this hypothesis by our empirical observations in Section~\ref{sec:observations} on existing instruction-tuned LLMs~\cite{InstructGPT,flan,instruction-tuning-survey,llama,llama2}~\footnote{For simplicity, we use LLMs to refer to instruction-tuned LLMs for the rest of the paper.} which have demonstrated that hidden states corresponding to the generated answers show considerably better instruction-awareness compared to those derived from the prompt.

Our observations indicate that function words and phrases in the answers do not contribute to better embedding quality. For instance, the introductory phrase ``\emph{Sure! Based on the input provided...}'' is irrelevant to the answers and is commonly found across various inputs. This redundancy can lead to inefficiency due to an increased decoding length, emphasizing the importance of answer brevity.

To effectively instantiate the embed-via-answering idea, 
we propose an \textbf{In}struction-following Em\textbf{bedder} framework (\model), which is compatible with both large language models (LLMs) and smaller encoder-based LMs such as RoBERTa.
Specifically, \model fine-tunes the LM on a union of $11$ abstractive question answering (QA) datasets with $\sim 200,000$ paragraph-question-answer triplets where the answers are usually short and informative. 
To facilitate the model to learn (implicit) semantics, we choose abstractive QA in particular, as the answers cannot be directly extracted.
We further simplify the answers by removing all the stopwords, resulting in an average answer length of $2.89$.

Due to the scarcity of evaluations focusing on instruction-following capabilities in the literature, we introduce a suite of tasks aimed at testing the ability of embedders to be instruction-aware, including
(1) a triplet task that selects the closer sentence to the anchor sentence based on two different instructions, 
(2) an instruction-following sentence similarity task, and 
(3) a task for clustering the same corpus under various instructions. 
Furthermore, we evaluate \model s' robustness to the instructions by testing it on clustering datasets with either correct, implicit, or incorrect instructions.
Our model is compared with both traditional text embedders as well as LLM-based embedders. 
The results demonstrate that our model can effectively process user instructions while generating high-quality embeddings.
Moreover, we empirically observe that the hidden states corresponding to the first generated token can already effectively follow instructions, {which makes it as efficient as traditional embedder methods} by only requiring one forward pass of the LM.
Finally, we propose to interpret the embedding clusters via post-processing on the generations of \model, and we observe that the clusters can reflect instruction-following capability when applying multiple instructions to the same corpus.

Our contributions are the following:
\begin{itemize}[nosep,leftmargin=*]
    \item We address a novel and challenging problem: instruction following of text embeddings and propose a framework, \model, to handle it by learning to answer user questions given inputs. 
    \item We provide a comprehensive assessment for instruction-following text embedders, including instruction awareness tests and instruction robustness tests, which intuitively reflect the models' instruction-following capability.
    \item We propose an approach for extracting explanations from embedding clusters. We show that these explanations further reflect instruction-following capability.
    \item We open source our code, datasets, and model checkpoints to facilitate future research: \url{https://github.com/zhang-yu-wei/InBedder}.
\end{itemize}

%% file: 2-related_work.tex
\subsection{Text Embedder}

Text embedders empower modern natural language processing systems with a wide variety of abilities like clustering~\cite{clustering_survey} and information retrieval \cite{DPR}. 
In the representation space of text embedders, similar texts are embedded close to each other. 
Thus, Siamese networks~\cite{sentence-bert} and contrastive learning~\cite{SimCSE} are proposed to learn the relative position of texts in the latent embedding space. 
Text embedders are further strengthened by incorporating more weakly supervised text similarity annotations~\cite{E5,BGE} and model structure variants \cite{GTR}. 
However, these mainstream text embedders only process general textual similarity, ignoring the changing view on textual similarity based on user demands. 
Recently, Instructor \cite{instructor} explores an instruction-based text embedder by concatenating instructions before the input texts. 
Our \model shows a substantially stronger instruction-following text embedder in instruction following by using expected answer distributions instead of concatenated instruction-text pairs as the representation.

\subsection{Instruction Tuning}

Instruction-following~\cite{instruction-tuning-survey} of LLMs is one of the core abilities for them to capture the user intents, which makes LLMs popular among users. 
InstructGPT~\cite{InstructGPT} is a first trial on instruction-following LLMs, which unearths the potential of LLMs to complete tasks under instructions from users. 
With an outstanding instruction-following ability from reinforcement learning with human feedback (RLHF), ChatGPT~\cite{chatgpt} achieves great success inside or outside the natural language community. 
The open-source instruction-following LLMs, like LLaMA~\cite{llama,llama2}, also provide valuable resources for researchers to study the instruction-following abilities of LLMs. 
Our study extends the idea of instruction-following from language modeling to text embeddings. 
Previously, it has been discovered that LLM can explore and manipulate various attributes of texts \cite{cotam}. Moreover, LLM hidden states can effectively represent space and time~\cite{gurnee2023language}, an aspect of texts such as honesty~\cite{zou2023representation} or a task defined by input-output pairs~\cite{todd2023function}. 
Despite the potential, it is still unknown how to effectively aggregate these hidden states to produce a high-quality representation.

\subsection{Goal-Driven Clustering}

With the recent advancements of instruction-following LLMs, goal-driven clustering has been proposed to group text corpora according to a personalized goal~\cite{goalex}. In order to address such a challenging yet novel problem, Goal-EX~\cite{goalex} applies a two-step pipeline that first proposes cluster explanations with GPT-4 and then selects clustering assignments with another LLM. A user-oriented goal is included in the proposed step. \citet{zhang-etal-2023-clusterllm} proposes another method that can incorporate user instructions to first determine sentence relationships via a triplet selection task and then produce clusters via fine-tuning. These produced clusters can be helpful for multi-document summarization~\cite{coavoux-etal-2019-unsupervised,fabbri-etal-2019-multi,lu-etal-2020-multi-xscience}, especially those that are personalized. Our paper on the other hand directly produces embeddings that are shaped according to different instructions applied which does not require calling APIs of LLMs and potentially saves costs.


%% file: 3-preliminary-yuwei.tex
\subsection{Instruction-following Embedder}

We introduce the definition of instruction-following embedder in this section.
A vanilla text embedder (denoted $\mathbf{Emb}(\cdot): \mathcal{X}\rightarrow \mathcal{Z}$)~\cite{reimers-2019-sentence-bert,SimCSE,ni-etal-2022-sentence,GTR,E5,BGE} embeds texts from token sequence space $\mathcal{X}$ into a $D$-dimensional vector space $\mathcal{Z}\subseteq \mathbb{R}^D$, where similarities between two pieces of texts can be measured by a certain metric $Sim(\cdot,\cdot): \mathcal{Z}\times\mathcal{Z} \mapsto \mathbb{R}$.
These embeddings are usually designed to generically represent texts, i.e., they aim to capture the overall meaning. 
Such an approach, while versatile, often fails to align with a specific downstream application,
e.g., grouping a corpus according to a particular interest or customizing a search engine with a targeted aspect.
In this paper, we assume these goals can be specified by a user instruction $I$ and then used to shape the embedding space without any fine-tuning to the text embedder. 
Under this circumstance, the similarity scores are conditional, i.e., $Sim(\cdot,\cdot|I)$.

The most straightforward approach is to just embed the concatenated instruction and input, which we will hereafter refer to as \textbf{prompt},
\begin{equation*}
\centering
\small
\begin{aligned}
Sim(X, X'|I) & = Sim(\mathbf{Emb}(I \oplus X), \mathbf{Emb}(I \oplus X'))  \\
\end{aligned}
\end{equation*}
where $X,X'\in \mathcal{X}$ and $\oplus$ is concatenation.
In order to assess the instruction-following ability, we will present a series of tasks in Section~\ref{sec:instruction_awareness} that require the model to understand the instructions.

\noindent \textbf{Instructor}~\cite{instructor}, a previous work, utilized a contrastive objective alongside multi-task learning to develop a more general text embedder.
Our experiments in Section~\ref{sec:instruction_awareness} demonstrate that their model does not adequately comprehend instructions.
This is not surprising given the limited instruction diversity and the lack of encouragement to follow instructions during training.

\noindent \textbf{Our hypothesis}.
We hypothesize the responses of LLMs~\cite{llama,llama2,chatgpt,flan} can be embedded to produce instruction-following embedding.
Specifically, the LLMs are prompted to generate a response $Y$,
\begin{equation*}
    \centering
    \small
    \begin{aligned}
        Y = \mathbf{LLM}(I \oplus X) \\
    \end{aligned}
\end{equation*}
where both $X,Y\in \mathcal{X}$ are from token sequence space. $\mathbf{LLM}(\cdot):\mathcal{X}\rightarrow\mathcal{X}$ is a function that maps prompts to responses. Usually, there could be multiple valid $Y$ for a given prompt.
In order to accommodate instruction-following embedding, we offer a novel viewpoint, which treats the instruction $I$ as a \textbf{question} about the input text $X$ and encodes the \textbf{expected answers} (i.e. the responses to the question).
In this paper, we study how to effectively embed expected answers.


\subsection{Instruction Awareness Tests}
Traditional generic embedding evaluation benchmarks, such as MTEB~\cite{muennighoff-etal-2023-mteb} and SentEval~\cite{conneau-kiela-2018-senteval}, lack the ability to assess instruction awareness. In this work, we propose a set of new tasks specifically designed to comprehensively evaluate the capabilities of embedding models in this regard.
We discuss the task formulations below and leave the detailed dataset creation procedures in Appendix~\ref{sec:task_create}.

\noindent \textbf{IntentEmotion.}
Inspired by previous works~\cite{zhang-etal-2023-clusterllm}, we employ triplet tasks with two contrasting criteria, \textit{i.e.} the intent and emotion of an utterance. A triplet task is composed of three different utterances $\{u_1,u_2,u_3\}$ where $u_1,u_2$ have the same intent but different emotions while $u_1,u_3$ have the same emotion but different intents, or vice versa. A success is defined under criterion $I^{int}$ if $d(z_1^{int},z_2^{int})<d(z_1^{int},z_3^{int})$. On the other hand, it is said to be a success for criterion $I^{emo}$ if $d(z_1^{emo},z_2^{emo})>d(z_1^{emo},z_3^{emo})$. Notice that the ranking is reverted under the two criteria. We use the harmonic mean of two success rates as our metric.

\noindent \textbf{InstructSTSB.}
Traditional Semantic Textual Similarity (STS) Benchmark~\cite{cer-etal-2017-semeval} lacks a definitive criterion for annotators to rely on, resulting in the subjectivity of the ratings. Hence, we create another instruction-based STS task where the two sentences are similar or dissimilar based on different instructions. We measure the Spearman correlation from cosine similarities. Notice that a similar dataset was first proposed in \citet{deshpande-etal-2023-c}. The main differences are that (1) our dataset is created directly from the original test set of STSB~\footnote{\url{https://huggingface.co/datasets/mteb/stsbenchmark-sts/viewer/default/test}} via brainstorming instructions; (2) our dataset only involves two ratings $0$ and $1$ indicating same or different, unlike the $1\sim 5$ rating scale in their case, reducing subjectivity in the evaluation.

\noindent \textbf{NYTClustering.}
We present the clustering results for the New York Times (NYT) dataset~\cite{sandhaus2008new}, which is categorized according to two annotations: topic and location of the news articles. The results are reported using the harmonic mean of the V-measure for both clustering types.

\subsection{Instruction Robustness Tests}
We further introduce an evaluation task specifically designed to assess the robustness of embedding models to various instructions. We employ clustering tasks to evaluate model performance in response to correct, implicit, and incorrect instructions. For each clustering task, a set of 10 correct instructions is generated by instructing GPT-4 to paraphrase the original task instructions. Similarly, a set of 10 implicit instructions is produced by GPT-4 through the rephrasing of the instructions to convey them implicitly. Moreover, 10 incorrect instructions are created by prompting GPT-4 to formulate instructions that diverge from the original task objective. Examples of these instructions are illustrated in Figure~\ref{fig:prompt_robustness_example}. The difference in average performance between correct and incorrect instructions is denoted as $\Delta_{ci}$, and the difference in average performance between implicit and incorrect instructions is denoted as $\Delta_{ii}$. See Appendix~\ref{sec:instruction_robustness_tests_create} for details.

%% file: 4-method-yuwei.tex
In this section, we introduce \model that is derived from observations on LLMs. We first define several ways to acquire sentence embeddings from LLMs in Section~\ref{sec:direct_vs_re}. Subsequently, we illustrate early observations in Section~\ref{sec:observations}. Finally, we introduce a framework that fine-tunes an LLM to an instruction-following embedder, \model, in Section~\ref{sec:inbedder}.

\subsection{Encoding Methods}\label{sec:direct_vs_re}
Contemporary LLMs are usually composed of one (encoder-/decoder-only) or two (encoder-decoder) transformer architectures with $L$ layers.
The input of the transformer is a sequence of embeddings $[h_0^1,\cdots,h_0^N]$ where $N$ is the length of prompt ($I\oplus X$). Each layer will then produce an intermediate hidden state $\mathbf{h}_l$ until the last layer which is used to predict the $(N+1)$th output token.
We first introduce two strategies to acquire a single aggregated embedding from an off-the-shelf LLM.

\noindent \textbf{Direct Encoding} directly utilizes LLM hidden states. Since it is not obvious which hidden states contain the most relevant information to the prompt, we explore $5$ aggregation methods for each layer:

\noindent 1)
The average of generation $Y$'s hidden states with generation length $N_g$, 
\begin{equation*}
    \mathbf{Emb}^{\text{\texttt{avg-gen}}}_l=\frac{1}{N_g + 1} \sum_{j=0}^{N_g} h_l^{(N+j)},
\end{equation*}
\noindent 2)
The average of prompt hidden states. This will serve as a direct comparison to ``\texttt{avg-gen}''.
\begin{equation*}
    \mathbf{Emb}^{\text{\texttt{avg-ppt}}}_l=\frac{1}{N-1} \sum_{i=1}^{N-1} h_l^i,
\end{equation*}
\noindent 3) The hidden states used to predict the first token in generations,
\begin{equation*}
    \mathbf{Emb}^{\text{\texttt{1st-gen}}}_l=h_l^N,
\end{equation*}
\noindent 4) The last generation hidden states,
\begin{equation*}
    \mathbf{Emb}^{\text{\texttt{last-gen}}}_l=h_l^{N+N_g},
\end{equation*}
\noindent 5) The average of all hidden states,
\begin{equation*}
    \mathbf{Emb}^{\text{\texttt{avg-all}}}_l=\frac{1}{N+N_g} (\sum_{i=1}^N h_l^i + \sum_{j=1}^{N_g} h_l^{N+j}),
\end{equation*}
In practice, we adjust the aggregation methods with regard to the uniqueness of each architecture. See details in Appendix~\ref{sec:details_direct_encoding}.

While direct encoding is commonly applied for conventional encoders \cite{E5,instructor}, using only the input information might not reveal the implicit features that can be inducted by answering the prompt. Thus, we propose \noindent \textbf{Re-encoding}, which is a two-step approach that first produces the responses $Y$ based on the prompts and then re-encode them using another embedder $\mathbf{Emb}_R$. Mathematically,
\begin{equation*}
    \mathbf{Emb}^{\text{re-enc}} = \mathbb{E}_{P(Y|I\oplus X)}[\mathbf{Emb}_R(Y)]
\end{equation*}
We then re-write the above with an empirical estimation,
\begin{equation*}
    \mathbf{Emb}^{\text{re-enc}} = \frac{1}{|\mathcal{S}_Y|} \sum_{{\tiny Y\sim \mathcal{S}_Y}}\mathbf{Emb}_R(Y)
\end{equation*}
where $\mathcal{S}_Y$ is sampled from response distribution $P(Y|I\oplus X)$. 
We choose $\mathbf{Emb}_R$ to be a (relatively) light-weight sentence transformer, thus the efficiency of re-encoding is similar to that of \texttt{avg-gen}.
And when $N_g=1$, all the aggregation methods possess the same efficiency.

\input{tables/observation}

\subsection{Answer Speaks Louder}\label{sec:observations}
In this section, we show some early observations that guide us towards the design of {\model}.
With the definitions in the previous section, we show the performance comparison among various aggregation methods on an existing LLM in Figure~\ref{fig:observation} left. To assess performance, we devised three distinct tasks that require the embedders to comprehend not only the raw texts but also the instructions, which will be further elaborated in Section~\ref{sec:instruction_awareness}. It is evident that hidden states derived from generations (\texttt{avg-gen}) consistently surpass those from prompts (\texttt{avg-ppt}). Additionally, averaging all hidden states, denoted as \texttt{avg-all}, does not enhance performance. Finally, an examination of three distinct models in Figure~\ref{fig:observation} reveals that re-encoding consistently outperforms all direct encoding methods, while increasing the sample size $|\mathcal{S}_Y|$ will further boost performance. These observations manifest our hypothesis that answers are more important for instruction-following embedder, in other words ``answers speak louder''.

\input{tables/direct_vs_re_barplot}

\subsection{Answer Brevity Matters}
One notable issue for using LLMs as embedders is their propensity to produce content that, while enhancing readability for humans, may not be directly relevant to the task at hand. For instance, \texttt{llama-2-7b-chat} frequently initiates responses with introductory phrases such as ``Based on the input provided...'' or ``The topic of the news article is...'' which are common across various requests. It is thus plausible to conjecture that the hidden states responsible for generating these superfluous contents contribute no useful information to the embedding task. Following this intuition, we conducted a simple experiment to validate the impact of filtering out hidden states associated with such content. Specifically, we compiled a list of candidate tokens for exclusion, which includes tokens present in the instruction, stopwords, and common phrases like ``Based on''. While calculating ``\texttt{avg-gen}'', we disregard hidden states linked to the generation of tokens from this list. The outcomes, depicted as green bars in Figure~\ref{fig:direct-vs-re}, indicate a marginal improvement in the performance of the three evaluated models upon the removal of non-informative content, thereby validating the assumption that these hidden states are indeed redundant.

\subsection{Our \model}\label{sec:inbedder}
In order to effectively instantiate the embed-via-answering, we propose a novel fine-tuning framework leveraging existing curated question-answering (QA) datasets. Specifically, we collect a set of $11$ abstractive QA datasets~\footnote{We also include several multiple-choice QA datasets but remove all the wrong choices.}, which sum up to $\sim200,000$ paragraph-question-answer triplets. As discussed in Section~\ref{sec:preliminary}, we treat the paragraph as the input, the question as the instruction, and generate the answers. Note that, we pre-process the answers so that all the stopwords are removed, which results in an average response length of $2.89$. As will be demonstrated in the experiments, such a pre-processing step significantly contributes to our method. We then fine-tune the LM with an autoregressive objective.

\begin{figure}[H]
    \centering
    \includegraphics[width=0.48\textwidth]{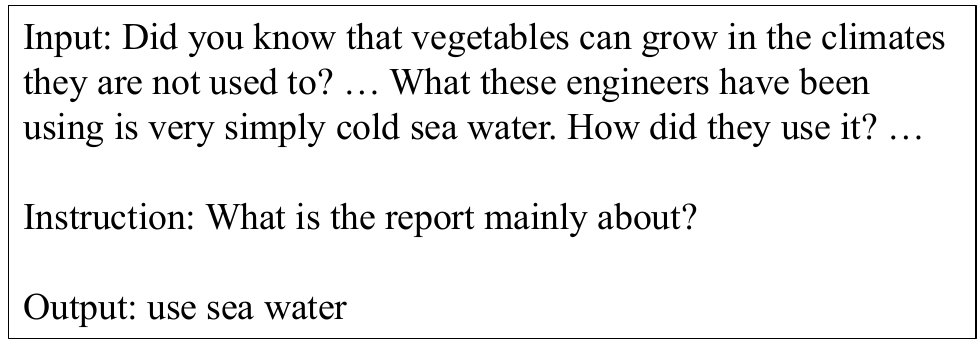}
    \caption{An example from our training data.}
    \label{fig:qa_example}
    \vspace{-3mm}
\end{figure}

We emphasize three inherent advantages of \model: (1) QA datasets usually have concise outputs that will promote the LLMs to respond eagerly without considering too much about readability. Refer to Figure~\ref{fig:qa_example} for an example. (2) Compared to multi-task datasets introduced in \citet{instructor}, our dataset offers significantly greater diversity in instructions, attributed to the variety of questions associated with each input paragraph, unconstrained by question format. And most importantly, the datasets are publicly available without any extra costs. (3) The auto-regressive objective induces better interpretability of generated embeddings via mining explanations from its generations.

%% file: tables/observation.tex
\begin{figure}[t]
    \centering
    \includegraphics[width=\linewidth]{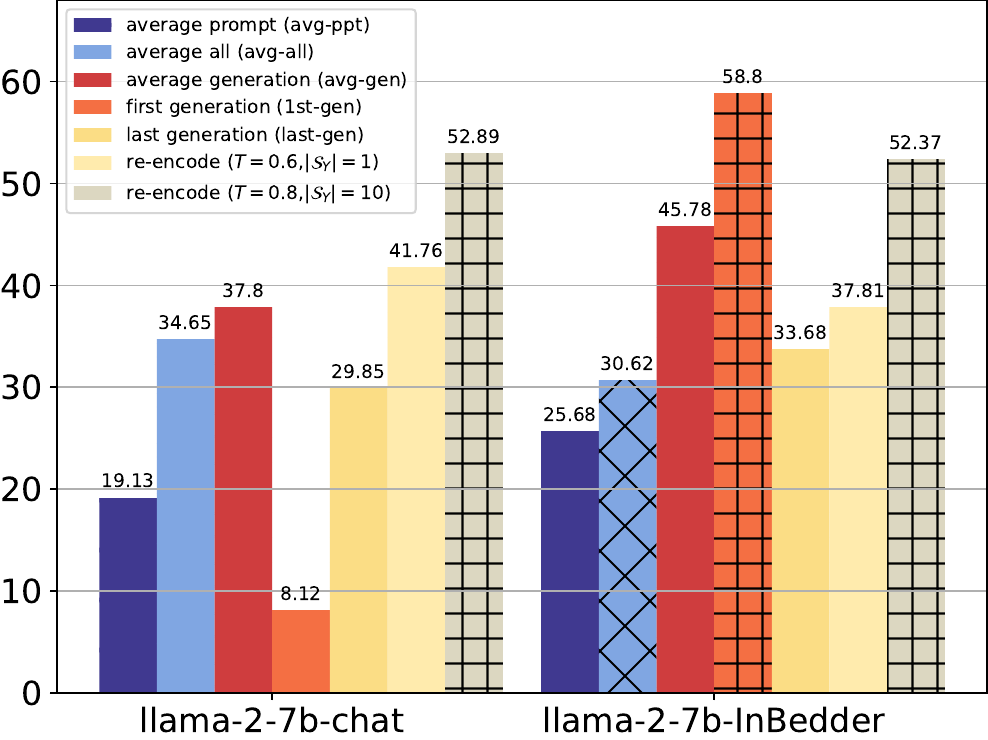}
    \caption{Instruction awareness tests performance (averaged over 3 datasets) for different encoding methods introduced in Section~\ref{sec:direct_vs_re} from the last layer. We show two models here \texttt{llama-2-7b-chat} from Huggingface and \texttt{llama-2-7b-\model} that is our fine-tuned model from \texttt{llama-2-7b}. $T$ is the decoding temperature while $\mathcal{S}_Y$ is the sample size. \textbf{Observations:} (1) The generation/answer side (i.e., the checkerboard pattern) is more informative than the prompt side (i.e., the dark blue with dotted pattern); and (2) In \texttt{llama-2-7b-\model}, \texttt{1st-gen} seems to significantly outperform others. See analysis of model depth in Figure~\ref{fig:model_depth}.}
    \label{fig:observation}
    \vspace{-2mm}
\end{figure}

%% file: tables/direct_vs_re_barplot.tex
\begin{figure}[t]
    \centering
    \includegraphics[width=0.45\textwidth]{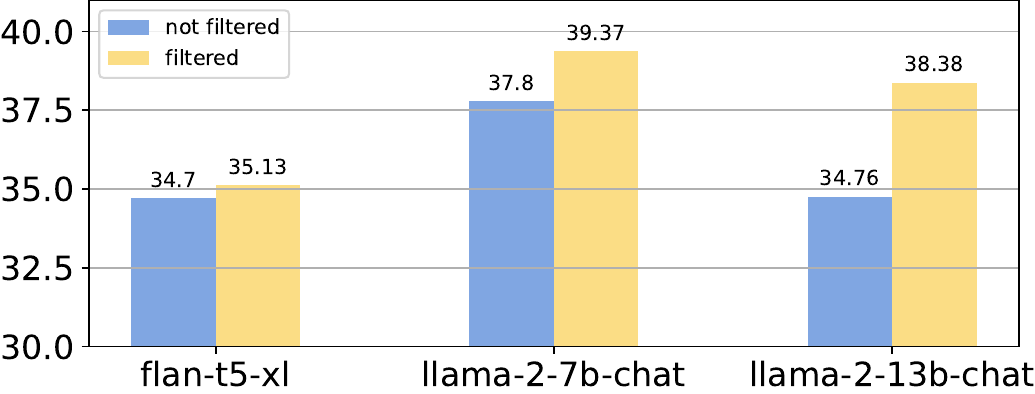}
    \caption{Filtered vs. not filtered (i.e., \texttt{avg-gen} on the last layer of each LLM). \textbf{Observations:} filtering hidden states associated with uninformative contents can marginally improve performance.}
    \label{fig:direct-vs-re}
    \vspace{-6mm}
\end{figure}

%% file: 5-experiment.tex

We introduce experimental setup in Section~\ref{sec:implementations} and Section~\ref{sec:compared_methods}. We then present results on proposed instruction awareness tests and instruction robustness tests in Section~\ref{sec:instruction_awareness} and Section~\ref{sec:prompt_robustness_tests}. Finally, we show a comparison of generic embedding tasks in Section~\ref{sec:generic_sentence_embedding_tasks}.

\subsection{Implementations}\label{sec:implementations}
We fine-tune \model from various kinds of language models such as (1) \texttt{roberta-large}, (2) \texttt{opt-1.3b}, (3) \texttt{opt-2.7b}, and (4) \texttt{llama-2-7b}~\footnote{huggingface ids: ``roberta-large'', ``facebook/opt-1.3b'',\\``facebook/opt-2.7b'', ``meta-llama/Llama-2-7b-hf''.}.
For masked language modeling-based roberta-large, we adapt our framework by appending mask tokens behind prompts with the same length as target tokens and then training with mask token prediction loss. During testing, we append $3$ mask tokens to represent the answer. We consistently train for $1$ epoch with a learning rate of $2\times10^{-5}$. For \model, we always employ the same pattern to feed the inputs to the models, i.e. ``\#\#\# Input:\textbackslash n\{input\}\textbackslash n\textbackslash n\#\#\# Instruction:\textbackslash n\{instruction\}\textbackslash n\textbackslash n\#\#\# Response:''. For \texttt{llama-2} chat models, we provide an extra prefix to induce shorter answers: ``Your task is to give an answer according to the instruction and input. Please keep your answer short.''.

At test time, we allow the maximum generation length to be $40$ for \texttt{llama-2} chat models and $3$ for our \model. We exclude hidden states corresponding to special tokens. We consistently use \texttt{e5-large-v2}~\cite{E5} as our re-encoder. Lastly, we set the maximum prompt length to be $512$ (including instruction, input, and the words in the pattern). We train and evaluate these models with at most $4\times$A100 (PCIe).

\input{tables/instruct_awareness_table}
\input{tables/mteb_subset}

\subsection{Compared Methods}\label{sec:compared_methods}
We compare with generic sentence embedding models: E5~\cite{E5} and Instructor~\cite{instructor}~\footnote{huggingface ids: ``intfloat/e5-large-v2'',\\ ``hkunlp/instructor-large''}. We also compare with instruction-tuned models: \texttt{llama-2} chat models~\cite{llama2} that are fine-tuned with RLHF~\footnote{huggingface id: ``meta-llama/Llama-2-7b-chat-hf'' and ``meta-llama/Llama-2-13b-chat-hf''}. For \texttt{roberta-large} and \texttt{opt} models we compare with those checkpoints tuned on Alpaca~\cite{alpaca}. For Alpaca fine-tuning, we follow the dataset and hyperparameters of the original implementation.~\footnote{\url{https://github.com/tatsu-lab/stanford_alpaca/tree/main}}

\input{tables/prompt_robustness}

\subsection{Instruction Awareness Tests Results}\label{sec:instruction_awareness}

\input{tables/explanation}

In Figure~\ref{fig:observation} right, quite unexpectedly, we observe that using \texttt{1st-gen} in \model achieves the best performance and it outperforms the other encoding methods by a significant amount. We hypothesize that although \texttt{1st-gen} is utilized solely for decoding the first token in the generations, it may contain the most relevant information due to the model being trained on concise outputs. Further qualitative analysis in Table~\ref{tab:first_token} shows that the first generated tokens usually correspond to the answer. We then present comparisons across various models in Table~\ref{tab:instruction_awareness}. Fine-tuning \model appears to be effective across a range of model sizes, from the 355M model \texttt{roberta-large} to the 1.3/2.7b \texttt{OPT} and the 7b \texttt{llama-2}. We can also observe that without pre-processing, the performance will be significantly degraded on instruction-awareness according to \texttt{llama-2-w/o-process}, which further validates that conciseness of outputs is important.

\begin{figure*}[t]
    \centering
    \includegraphics[width=\textwidth]{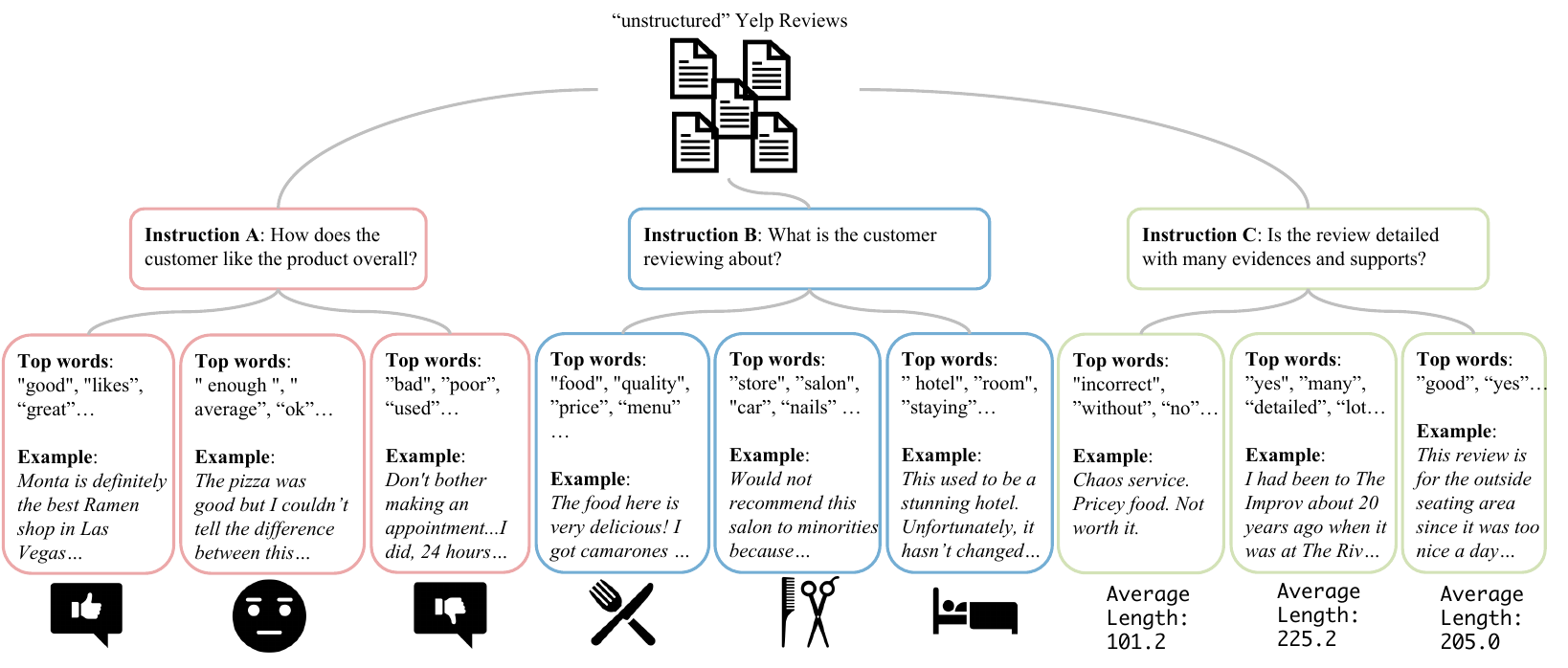}
    \caption{Instruction-following clustering with \model. The results are produced by simply instructing the model. $3$ clusters along with top words and examples are shown for each instruction where we can observe clear accountability to the instructions.}
    \label{fig:application}
\end{figure*}

\subsection{Instruction Robustness Tests Results}\label{sec:prompt_robustness_tests}
Figure~\ref{fig:prompt_robustness} presents the results obtained across three models, and see more datasets in Figure~\ref{fig:prompt_robustness_more}. Compared to \texttt{instructor-large} and \texttt{llama-2-7b-chat}, our model demonstrates larger values of $\Delta_{ci}$ and $\Delta_{ii}$, as well as superior average performance when applying correct instructions. This indicates that our model exhibits a better understanding of correct or implicit instructions and possesses greater robustness against incorrect instructions.

\subsection{Generic Sentence Embedding Tasks}\label{sec:generic_sentence_embedding_tasks}
Finally, we also compare performances on generic sentence embedding tasks. We choose a subset of tasks from the original MTEB~\cite{muennighoff-etal-2023-mteb} benchmark, including: ``TwentyNewsgroupsClustering'', ``AskUbuntuDupQuestions'', ``SciDocsReranking'' and ``StackOverflowDupQuestions''. The first task is a clustering task with V-measure as a metric. The last three tasks are reranking tasks that require the model to correctly identify the ones that are close to the query with cosine similarity metrics. We use the ``mean average precision (MAP)'' as our metric which is reported in MTEB. For each task, we design a task-level prompt that describes the requirements. We observe in Table~\ref{tab:mteb_subset} that our \model has a close performance to state-of-the-art embedders E5~\cite{E5} and Instructor~\cite{instructor} than other LLM-based embedders, even though it was not trained with a contrastive objective as in the state-of-the-art sentence transformers.

\section{Embedder Clustering Interpretation}
Interpreting neural embeddings has long been an aspiration in numerous research endeavors~\cite{panigrahi-etal-2019-word2sense,trifonov-etal-2018-learning}. We show in this section that \model naturally possesses interpretability due to its instruction-following training objective. Specifically, we propose a method to ``extract answers'' from semantic clusters produced by the embedder.

\subsection{Interpretation Methods}
We directly post-process the generated sequences of \model to collect identifiable information about a cluster. To differentiate clusters, we initially collect outputs from each cluster following $K$-means clustering and concatenate these outputs into a single document per cluster. Subsequently, we employ Term Frequency-Inverse Document Frequency (Tf-idf) to vectorize these $K$ documents, resulting in $K$ feature vectors. The dimensions of each vector denote the relative frequency of a word's occurrence in one document compared to its occurrence in others. Hence, we rank feature words according to the corresponding value in the feature vector, which will then be designated as cluster keywords. 


\subsection{Results}
Table~\ref{tab:explanation} presents explanations derived from \texttt{llama-2-7b-\model}. When compared to the label components of each cluster, the top words collected effectively capture the unique characteristics of each cluster. To showcase the instruction-following capability of \model, cluster explanations are further illustrated with three distinct instructions in Figure~\ref{fig:application} using the Yelp review dataset~\cite{yelpreview} (originally designed for sentiment analysis). The top words distinctly delineate the differences between clusters, in accordance with the provided instructions. For example, under ``Instruction B'' various different products that the customers are reviewing about are produced from clusters, while on the other hand, under ``Instruction C'', variations in average sentence lengths are observed, indicating the degree of detail present in the review.

%% file: tables/instruct_awareness_table.tex
\begin{table}[t]
\centering
\scalebox{0.71}{
\begin{tabular}{lcccc}
\toprule
Model                                            & \multicolumn{1}{l}{I.STSB} & \multicolumn{1}{l}{IntEmo} & \multicolumn{1}{l}{NYT} & \multicolumn{1}{l}{Avg} \\
\midrule
\small e5-large-v2(w/o instruction)                                     & 0.00                         & 30.24                      & 50.07                   & 26.77                   \\
\small instructor-large                       & -15.02                        & 47.96                      & 49.96                   & 27.63                   \\
\midrule
\small roberta-large-alpaca(avg-gen)                    & 8.43                         & 90.34                      & 21.60                   & 40.12                   \\
\small roberta-large-\model(avg-gen) & 14.81                        & \bf 91.07                      & 51.18                   & 52.35                   \\
\midrule
\small opt-1.3b-alpaca(avg-gen)                         & -1.81                        & 71.51                      & 12.88                   & 27.53                   \\
\small opt-1.3b-\model(1st-gen)      & 7.47                         & 89.96                      & 53.13                   & 50.19                   \\
\midrule
\small opt-2.7b-alpaca(avg-gen)                         & 3.95                         & 75.09                      & 13.32                   & 30.79                   \\
\small opt-2.7b-\model(1st-gen)      & 10.45                        & 84.54                      & 59.43                   & 51.47                   \\
\midrule
\small llama-2-7b-chat(re-enc)                          & 16.56                        & 79.32                      & 29.41                   & 41.76                   \\
\small llama-2-13b-chat(re-enc)                         & 19.76                        & 73.60                      & 32.74                   & 42.03                   \\
\small llama-2-7b-w/o-process(1st-gen)                       & 21.10                        & 83.64                      & 52.72                   & 52.49                   \\
\small llama-2-7b-\model(1st-gen)    & \bf 22.07                        & 89.68                      & \bf 64.65                   & \bf 58.80   \\
\bottomrule
\end{tabular}
}
\vspace{-3mm}
\caption{Instruction awareness tests results. The best encoding methods are shown in parentheses for each non-sentence-transformer model. We only consider the last layer in this table. I.STSB is short for InstructSTSB.}
\label{tab:instruction_awareness}
\end{table}

%% file: tables/mteb_subset.tex
\begin{table}[t]
\centering
\scalebox{0.63}{
\begin{tabular}{lccccc}
\toprule
Model                                                                & AskU. & SciD. & StackO. & 20news & Avg \\
\midrule
\small e5-large-v2(w/o instruction)                                  & 59.01                         & {83.84}                & {50.60}                  & 47.94                          & 60.35                       \\
\small instructor-large                             & {63.48}                & 81.83                         & 50.50                           & {53.51}                 & 62.33                       \\
\midrule
\small roberta-large-alpaca(avg-gen)                & 56.29                         & 73.02                         & 41.66                           & 40.61                          & 52.90                       \\
\small roberta-large-\model(avg-gen) & 55.50                         & 73.80                         & 41.00                           & 41.93                          & 53.06                       \\
\midrule
\small opt-1.3b-alpaca(avg-gen)                     & 55.89                         & 69.68                         & 42.43                           & 38.49                          & 51.62                       \\
\small opt-1.3b-\model(1st-gen)      & 59.09                         & 71.33                         & 43.08                           & 46.45                          & 54.99                       \\
\midrule
\small opt-2.7b-alpaca(avg-gen)                     & 55.65                         & 76.26                         & 42.45                           & 32.11                          & 51.62                       \\
\small opt-2.7b-\model(1st-gen)      & 59.94                         & 75.33                         & 41.93                           & 49.07                          & 56.57                       \\
\midrule
\small llama-2-7b-chat(re-enc)                      & 55.26                         & 75.81                         & 41.43                           & 25.34                          & 49.46                       \\
\small llama-2-13b-chat(re-enc)                     & 53.69                         & 77.64                         & 38.84                           & 30.77                          & 50.24                       \\
\small llama-2-7b-w/o-process(1st-gen)                   & 61.25                         & 83.13                         & 44.39                           & 50.68                          & 59.86                       \\
\small llama-2-7b-\model(1st-gen)    & 60.32                         & 80.61                         & 44.77                           & 52.33                          & 59.51                      \\
\bottomrule
\end{tabular}
}
\vspace{-3mm}
\caption{Generic sentence embedding task performance. The best encoding methods are shown in parentheses for each non-sentence-transformer models. We only consider the last layer in this table.}
\label{tab:mteb_subset}
\end{table}

%% file: tables/prompt_robustness.tex
\begin{figure*}[t]
    \centering
    \begin{subfigure}[b]{0.32\textwidth}
        \includegraphics[width=\textwidth]{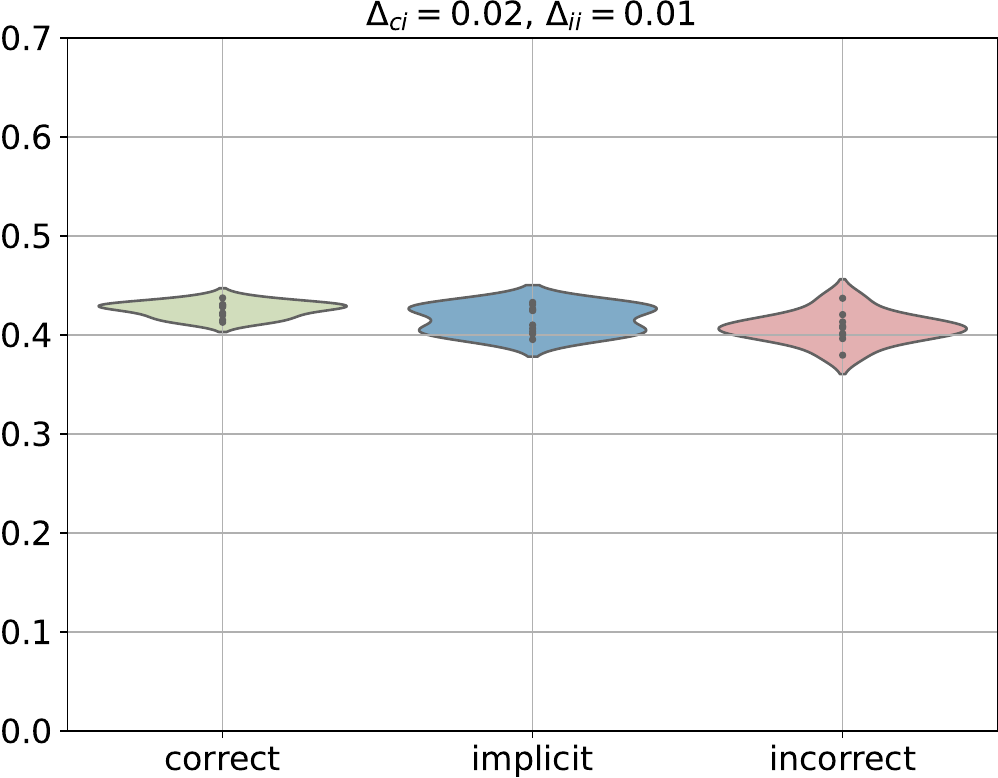}
        \caption*{(a) instructor-large}
    \end{subfigure}
    \begin{subfigure}[b]{0.32\textwidth}
        \subcaption*{FewNerd}
        \includegraphics[width=\textwidth]{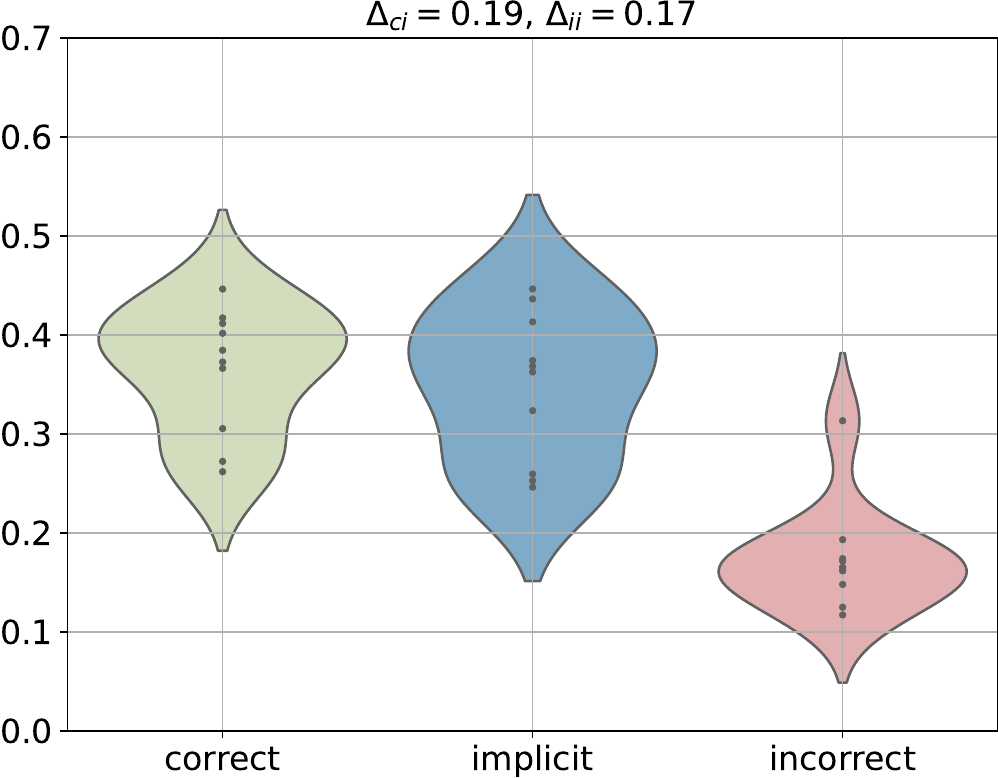}
        \caption*{(b) llama-2-7b-chat (re-enc)}
    \end{subfigure}
    \begin{subfigure}[b]{0.32\textwidth}
        \includegraphics[width=\textwidth]{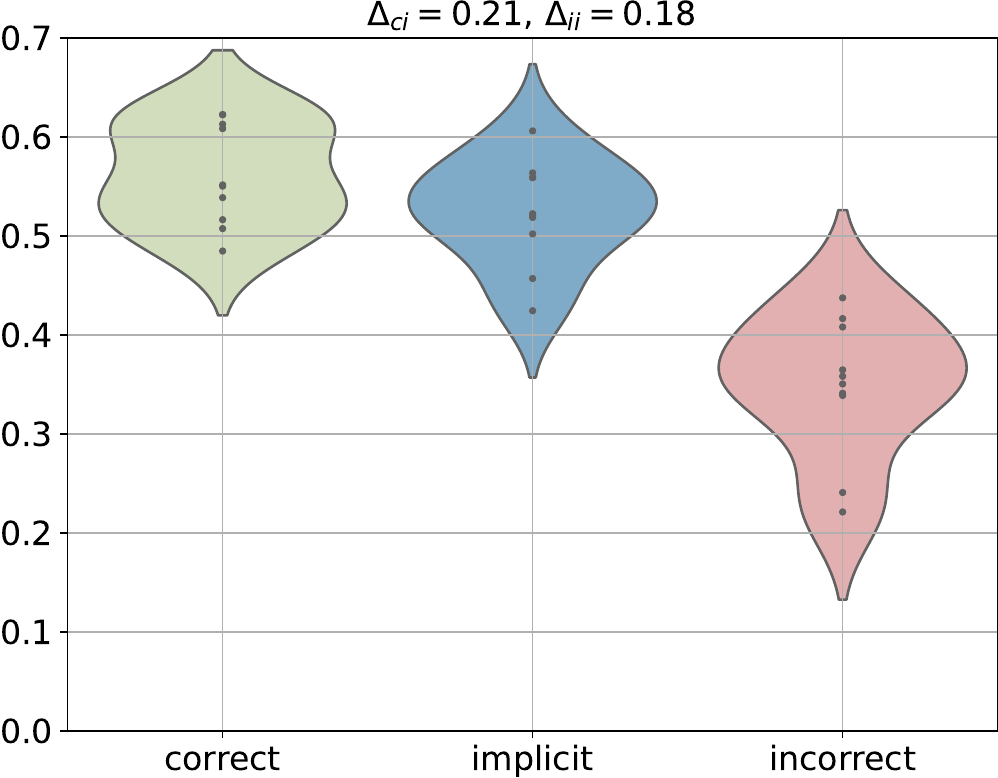}
        \caption*{(c) llama-2-7b-\model (1st-gen)}
    \end{subfigure}
    \caption{Instruction robustness tests results. Three set of instructions are tested: correct, implicit and incorrect. $\Delta_{ci}$ denotes the separation between mean of correct and incorrect. $\Delta_{ii}$ denotes the separation between mean of implicit and incorrect. \model shows better robustness and performance overall. See more datasets in Figure~\ref{fig:prompt_robustness_more}}
    \label{fig:prompt_robustness}
\end{figure*}

%% file: tables/explanation.tex
\begin{table*}[ht!]
\centering
\scalebox{0.85}{
\begin{tabular}{lllll}
\toprule
          & \multicolumn{4}{c}{RateMyProf}                                                                                                                                                                                                                                                                                                                                                                                                                                                                                                                    \\
\midrule
cluster   & 1 (lowest entropy)                                                                                                                 & 2                                                                                                                                    & 3                                                                                                                                  & 4 (highest entropy)                                                                                                                   \\
\midrule
components & \begin{tabular}[c]{@{}l@{}}personal qualities:327\\ ease or difficulty:31\\ assessment-related:15\\ amount of work:10\end{tabular} & \begin{tabular}[c]{@{}l@{}}amount of work:331\\ assessment-related:215\\ ease or difficulty:128\\ personal qualities:12\end{tabular} & \begin{tabular}[c]{@{}l@{}}assessment-related:173\\ personal qualities:99\\ ease or difficulty:77\\ amount of work:62\end{tabular} & \begin{tabular}[c]{@{}l@{}}ease or difficulty:338\\ amount of work:171\\ assessment-related:171\\ personal qualities:136\end{tabular} \\
\midrule
top words & \begin{tabular}[c]{@{}l@{}}teaching,personality,\\ classroom,quality,\\ good,teacher,\\ skills,student...\end{tabular}             & \begin{tabular}[c]{@{}l@{}}assignments,homework,\\ workload,expectations,\\ tests,difficulty,\\ professor,exams...\end{tabular}      & \begin{tabular}[c]{@{}l@{}}teaching,quality,\\ grade, ability,\\ style,grading,\\ lectures,enough...\end{tabular}                  & \begin{tabular}[c]{@{}l@{}}difficulty,professor\\ level,coursework\\ lectures,tests\\ class,course...\end{tabular} \\
\bottomrule
\end{tabular}
}
\caption{Cluster explanation results using generations from llama-2-7b-\model on RateMyProf (we show other datasets in Table~\ref{tab:explanation_app}). Clusters are ordered by increasing entropy, with entropy being determined by the distribution of labels within each cluster. Lower entropy indicates that the cluster's components are ``pure'' according to the labels. The table delineates the label components associated with each cluster, as indicated in the "components" row. Additionally, the top-$8$ words extracted by our interpretation method for each cluster are listed under "top words". Notice that we simplify some label names for presentation.}
\vspace{-3mm}
\label{tab:explanation}
\end{table*}

%% file: 6-conclusion.tex
Our work addresses a novel problem, text embedding with instruction-following. We propose \model to produce desirable embeddings from LLMs via generating expected answers. The method is inspired by observations on existing LLMs. Our text embedder model llama-2-7b-\model outperforms both traditional sentence transformers and aggregated embeddings from LLMs on instruction-awareness tests, and instruction robustness tests and achieves close performance on traditional generic tasks. We also show that \model inherently is applicable for embedding cluster explanation which will significantly facilitate user-oriented dataset analysis. We encourage future works to investigate more efficient solutions which is important in large-scale retrieval systems.

%% file: 7-limitations.tex
\noindent\textbf{Efficiency.}
Our model is not sufficiently efficient for large-scale retrieval tasks. In retrieval, corpus is usually encoded as vector embeddings beforehand, the only operation conducted is to encode the query and to compute the cosine similarities between the query and corpus. However, \model requires encoding the entire corpus \textit{w.r.t.} each user query which results in significant latency. However, one possible solution is to first select the most similar candidates and then use \model as a query-dependent reranker.

\noindent\textbf{Effectiveness on generic tasks.}
The results in Table~\ref{tab:mteb_subset} show that \model does not surpass traditional sentence transformers on especially generic reranking tasks. (1) Our ambition is to provide an instruction following embedder that could potentially facilitate user-oriented tasks rather than optimizing for high-performing sentence embedding and we leave the exploration on that dimension in future works. (2) \model might benefit from better prompt design or task description which we have discussed in Section~\ref{sec:prompt_robustness_tests}.

%% file: 8-appendix.tex
\section{Instruction Awareness Tests Creation}\label{sec:task_create}
\noindent \textbf{IntentEmotion}
We use BANKING77~\cite{banking77} test set as our base dataset to create triplets. We prompt \texttt{gpt-4-0613} to create utterances that have the same intent but two different emotions ``optimistic'' and ``frustrating'', denoted as $u_{opt}^1$ and $u_{fru}^1$, with the following prompt.
\begin{quote}
    \textit{Could you modify the emotion (one optimistic and one frustrating) of following utterance without changing the intent ("[INTENT]")? \\
    "[TEXT]" \\
    Please output a JSON object containing keys "optimistic" and "frustrating", and no other things.}
\end{quote}
For each generated utterance, we then prompt the same LLM again to modify the intent of the utterance, denoted as $u_{opt}^2$ and $u_{fru}^2$ with the following prompt.
\begin{quote}
    \textit{Modify the intent of the above utterances (i.e. from "[INTENT]" to another one that you brainstormed. Usually by modifying the objects or actions) without changing the emotions. Same as before, output a JSON object containing keys "optimistic" and "frustrating", and no other things.}
\end{quote}
This will result in $4$ generated utterances (disregarding the original utterance), then we group these utterances into $4$ triplets according to two criteria:
\begin{align*}
    \{u_{opt}^1,u_{opt}^2,u_{fru}^1\},\{u_{fru}^1,u_{fru}^2,u_{opt}^1\},\\\{u_{opt}^1,u_{fru}^1,u_{opt}^2\},\{u_{fru}^1,u_{opt}^1,u_{fru}^2\}
\end{align*}
In each triplet, the first one is the anchor, the second is the positive and the last is the negative. Thus the first two triplets follow emotion criterion while the last two follow intent criterion. As a result, there are $12,320$ triplets in total, half for emotion and half for intent. We calculate the triplet success rates for both criteria separately, and then calculate the harmonic mean.

\noindent \textbf{InstructSTSB}
We use STSb~\cite{cer-etal-2017-semeval} test set as out base dataset to generate sentence pairs. We generate two instructions, one that can discriminate the sentence pair and the other that can not. To achieve that, we prompt \texttt{gpt-4-1106-preview} sequentially with the following two instructions.
\begin{quote}
    The following two sentences have similar surface forms:\\

    1. [SENTENCE1]\\
    2. [SENTENCE2]\\
    
    In order to discriminate the two sentences, what question would you ask? (e.g. what is the subject of the sentence?) Please output a JSON object that contains the key "question".
\end{quote}
\begin{quote}
    Similar to the above, in order to make the answers to the two sentences immune to discrimination, what question would you ask? (e.g. what is the subject of the sentence?) Please output a JSON object that contains the key "question".
\end{quote}
As a result, there are 2758 sentence pairs in total. We then set the ratings for discriminative pairs to $0$ and $1$ for non-discriminative pairs. Following previous implementation~\cite{muennighoff-etal-2023-mteb}, we use spearman correlation as our metric and cosine similarity as similarity measurement.

\noindent \textbf{NYTClustering}
There are no further modifications to this dataset since it already contains two sets of annotations, one for location and one for topic.

\section{Instruction Robustness Tests Creation}\label{sec:instruction_robustness_tests_create}
We adopt clustering datasets FewNerd, FewRel and FewEvent from \citet{zhang-etal-2023-clusterllm}. We adapt clustering datasets RateMyProf and Feedbacks from \citet{goalex}. All these datasets are clustered under a complex task instruction such as entity type or the aspect of the review or the reason to (dis)like. Since the original paper~\cite{goalex} does not provide the annotations, we use \texttt{gpt-4-1106-preview} to select annotations for them and then we post-process the dataset so that the clusters are equal in size.
As a result, Feedbacks contains $3$ clusters and $756$ human feedbacks to machine generated data. RateMyProf contains $4$ clusters and $2,296$ reviews from RateMyProfessor.
Lastly, we provide various instructions that are correct, implicit or incorrect by prompting GPT-4 (webpage) to generate similar, implicit, or dissimilar instructions.

\section{Details on Direct Encoding}\label{sec:details_direct_encoding}
The direct encoding proposed in Section~\ref{sec:direct_vs_re} are all compatible with decoder-only transformers. For encoder-decoder models such as \texttt{flan-t5}, because of the two separated models, we remove \texttt{avg-all} since the hidden states are not in the same space. Besides, we extract \texttt{avg-ppt} from encoder and \texttt{avg-gen}\&\texttt{1st-gen}\&\texttt{last-gen} from decoder respectively. Notice that for \texttt{1st-gen}, we use the hidden states for the BOS token in the decoder side. For encoder-only models, we remove \texttt{1st-gen} and \texttt{last-gen}. We implement the sentence embedding function by generating tokens first~\footnote{Notice that, in huggingface~\cite{wolf2019huggingface}, both decoder-only and encoder-decoder model can use ``generation'' function: \url{https://huggingface.co/docs/transformers/main_classes/text_generation}. For encoder-only, we simply concatenate the ``[MASK]'' tokens after the prompts for generation.} and then cache the intermediate hidden states for further compute. Considering the efficiency, \texttt{avg-ppt}\&\texttt{1st-gen} only require single forward pass while the others require iterative generations and thus depending on the generation length.

\begin{figure*}[ht!]
    \centering
    \begin{subfigure}[b]{0.49\textwidth}
        \centering
        \includegraphics[width=\textwidth]{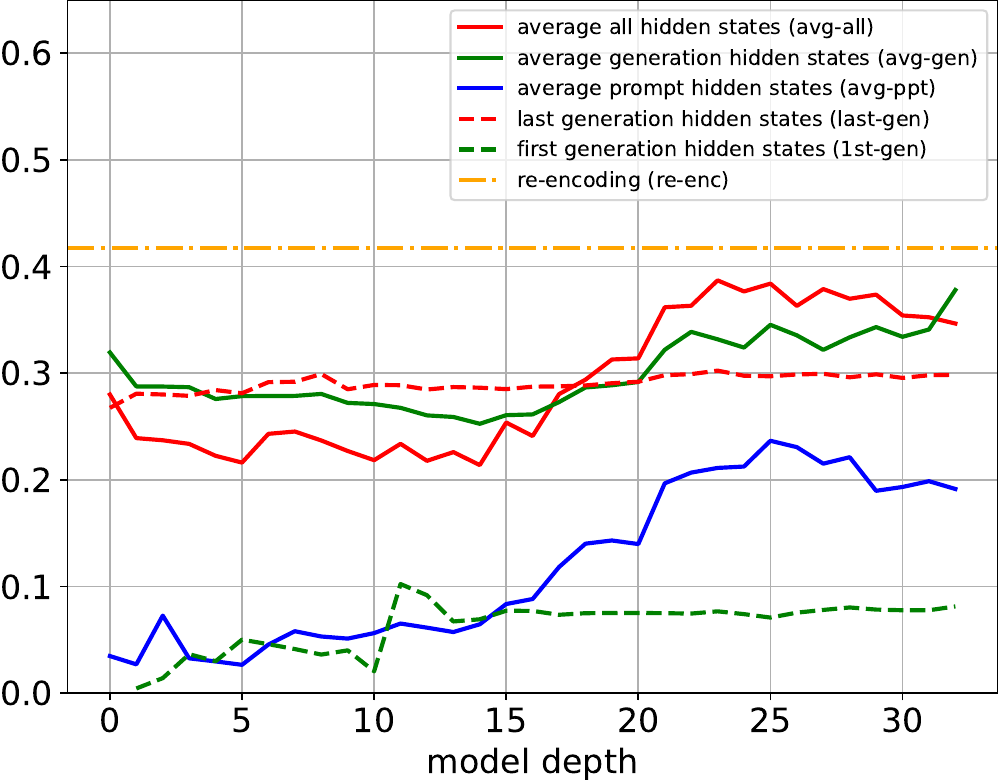}
        \caption{\texttt{llama-2-7b-chat}}
    \end{subfigure}
    \begin{subfigure}[b]{0.49\textwidth}
        \centering
        \includegraphics[width=\textwidth]{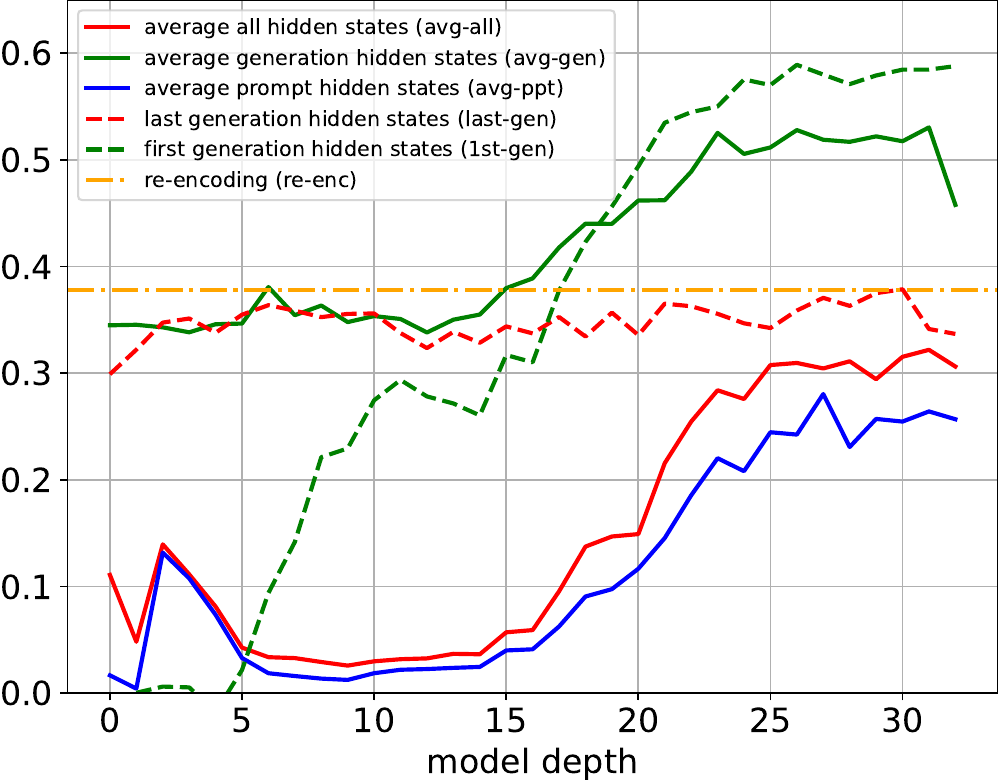}
        \caption{\texttt{llama-2-7b-\model}}
    \end{subfigure}
    \caption{Instruction awareness tests results vs. model depth.}
    \label{fig:model_depth}
\end{figure*}

\input{tables/explanation_appendix}

\input{tables/prompt_robustness_more}

\begin{figure}
    \centering
    \includegraphics[width=0.47\textwidth]{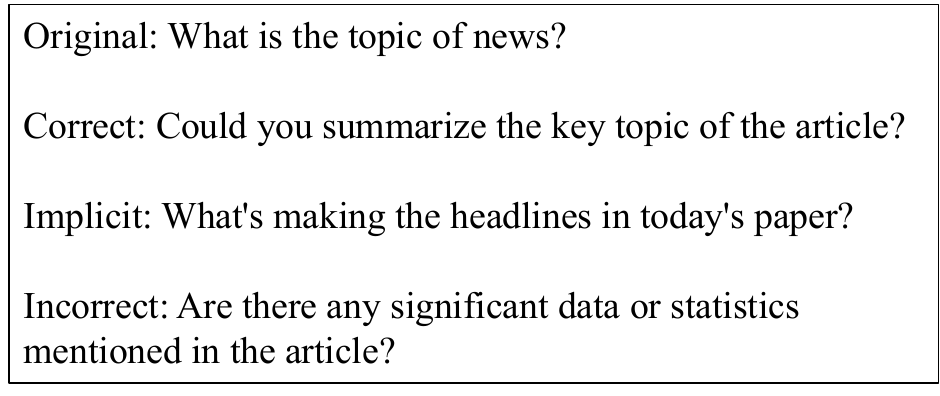}
    \caption{An example from our prompt robustness tests.}
    \label{fig:prompt_robustness_example}
\end{figure}

\begin{table*}[t]
\centering
\scalebox{0.53}{
\begin{tabular}{lll}
\toprule
Prompt                                                                                                                                                                                                                                                                                                                                                                                                                                                                    & top-$10$ first decoding                                                                      & GPT-4 answer                     \\
\midrule
\begin{tabular}[c]{@{}l@{}}\#\#\# Input:\\ The Justice Department filed suit Thursday against the state of Mississippi\\ for failing to end what federal officials call "disturbing" abuse of juveniles\\ and "unconscionable" conditions at two state-run facilities.\\ \\ \#\#\# Instruction:\\ What specific language or descriptors does the first sentence use to describe\\ the abuse and conditions at the juvenile facilities?\\ \\ \#\#\# Response:\end{tabular} & {[}'Dist', 'dist', 'des', 'D', 'Des', 'un', 'Un', 'use', 'uses', 'specific'{]}             & "disturbing," "unconscionable"   \\
\midrule
\begin{tabular}[c]{@{}l@{}}\#\#\# Input:\\ "Further testing is still under way, but at this stage, given the early detection,\\ the outlook in such instances would be positive," the specialist said yesterday.\\ \\ \#\#\# Instruction:\\ What additional information is provided in the first sentence that is not\\ present in the second sentence?\\ \\ \#\#\# Response:\end{tabular}                                                                                & {[}'fur', 'ear', 'testing', 'out', 'stage', 'first', 'F', 'd', 'special', 'information'{]} & Further testing, early detection \\
\midrule
\begin{tabular}[c]{@{}l@{}}\#\#\# Input:\\ Frank Quattrone, the former Credit Suisse First Boston technology\\ investment-banking guru, reportedly pleaded not guilty Tuesday to charges of\\ obstruction of justice and witness tampering.\\ \\ \#\#\# Instruction:\\ Who pleaded not guilty to charges of obstruction of justice and witness\\ tampering?\\ \\ \#\#\# Response:\end{tabular}                                                                            & {[}'Fran', 'former', 'Form', 'F', 'Cred', 'f', 'Qu', 'Mr', 'cred', 'ex'{]}                 & Frank Quattrone                 \\
\bottomrule
\end{tabular}
}
\caption{Qualitative analysis on the top-$10$ tokens decoded at the first position. We also present the answers from GPT-4 (webpage) by prompting it to ``answer this question within 5 words''.}\label{tab:first_token}
\end{table*}

%% file: tables/explanation_appendix.tex
\begin{table*}[ht!]
\centering
\scalebox{0.95}{
\begin{tabular}{llll}
\toprule
          & \multicolumn{3}{c}{Feedbacks} \\
\midrule
cluster   & 1 (lowest entropy)                                                                                                                                             & 2                                                                                                                                                                 & 3 (highest entropy)                                                                                                                                              \\
\midrule
compoents & \begin{tabular}[c]{@{}l@{}}Structure, Coherence:229\\ Inclusion of Main Points:9\\ Content Accuracy:7\end{tabular} & \begin{tabular}[c]{@{}l@{}}Content Accuracy:148\\ Inclusion of Main Points:134\\ Structure, Coherence:10\end{tabular} & \begin{tabular}[c]{@{}l@{}}Inclusion of Main Points:109\\ Content Accuracy:97\\ Structure, Coherence:13\end{tabular} \\
\midrule
top words & \begin{tabular}[c]{@{}l@{}}sentence,better,\\ structure,written,\\ improved,flow,\\ unclear,read...\end{tabular}                                               & \begin{tabular}[c]{@{}l@{}}dislike,accurate,\\ author,generated,\\ mention,machine,\\ accuracy,inaccurate...\end{tabular}                                         & \begin{tabular}[c]{@{}l@{}}like,feedback,\\ post,likes,\\ good,human,\\ advice,relationship...\end{tabular}            \\
\bottomrule
\end{tabular}
}
\\
\vspace{1mm}
\scalebox{0.65}{
\begin{tabular}{llllll}
          & \multicolumn{5}{c}{FewRel}                                                                                                                                                                                                                                                                                                                                                                                                                                                                                                                                                                                                                                  \\
\midrule
cluster   & 1 (lowest entropy)                                                                                               & 2                                                                                                                         & 3                                                                                                                           & 63                                                                                                                            & 64 (highest entropy)                                                                                                                           \\
\midrule
compoents & taxon rank:24                                                                                                    & \begin{tabular}[c]{@{}l@{}}heritage designation:70\\ located in the administrative...:1\\ location:1\end{tabular}         & \begin{tabular}[c]{@{}l@{}}taxon rank:46\\ instance of:1\\ said to be the same as:1\\ country of citizenship:1\end{tabular} & \begin{tabular}[c]{@{}l@{}}movement:9\\ religion:6\\ work location:6\\ said to be the same as:5\end{tabular}                  & \begin{tabular}[c]{@{}l@{}}language of work or name:23\\ said to be the same as:16\\ followed by: =10\\ applies to jurisdiction:7\end{tabular} \\
\midrule
top words & \begin{tabular}[c]{@{}l@{}}family,species,\\ gastropod,marine,\\ sea,urothoidae,\\ psolidae,feed...\end{tabular} & \begin{tabular}[c]{@{}l@{}}historic,registe,\\ places,listed,\\ national,historical,\\ house,significance...\end{tabular} & \begin{tabular}[c]{@{}l@{}}family,subfamily,\\ order,genus,\\ families,tribe,\\ sent,orders...\end{tabular}                 & \begin{tabular}[c]{@{}l@{}}friends,beowulf$^*$,\\ personalities,jewish,\\ slave,personality,\\ independence,owner...\end{tabular} & \begin{tabular}[c]{@{}l@{}}language,minor,\\ wikipedia,candela,\\ french,translation,\\ flag,major...\end{tabular}\\
\bottomrule
\end{tabular}
}
\\
\vspace{1mm}
\scalebox{0.72}{
\begin{tabular}{lllllll}
          & \multicolumn{6}{c}{FewNerd}                                                                                                                                                                                                                                                                                                                                                                                                                                                                                                                                                                                                                                                 \\
\midrule
cluster   & 1 (lowest entropy)                                                                                      & 2                                                                                                          & 3                                                                                                    & 56                                                                                                                  & 57                                                                                                       & 58 (highest entropy)                                                                                           \\
\midrule
components & \begin{tabular}[c]{@{}l@{}}Geo-Political:139\\ film:1\\ car:1\\ education: 1\end{tabular}               & \begin{tabular}[c]{@{}l@{}}Geo-Political:92\\ company:1\\ government:1\\ sports team:1\end{tabular}        & \begin{tabular}[c]{@{}l@{}}award:35\\ living thing:1\\ broadcast program:1\end{tabular}              & \begin{tabular}[c]{@{}l@{}}language:33\\ Geo-Political:8\\ written art:4\\ software:4\end{tabular}                  & \begin{tabular}[c]{@{}l@{}}artist, author:47\\ scholar:25\\ actor:9\\ Geo-Political:9\\ ...\end{tabular} & \begin{tabular}[c]{@{}l@{}}film:25\\ broadcast program:12\\ Geo-Political:9\\ written art:6\\ ...\end{tabular} \\
\midrule
top words & \begin{tabular}[c]{@{}l@{}}city,america,\\ europe,continent\\ usa,state,\\ north,county...\end{tabular} & \begin{tabular}[c]{@{}l@{}}country,europe,\\ jordan,ireland,\\ india,america,\\ uk,germany...\end{tabular} & \begin{tabular}[c]{@{}l@{}}award,prize,\\ awards,best,\\ given,show,\\ film,category...\end{tabular} & \begin{tabular}[c]{@{}l@{}}language,spoken\\ dialect,sentence\\ languages,skerry,\\ dialects,french...\end{tabular} & \begin{tabular}[c]{@{}l@{}}person,artist\\ dan,name\\ paris,well,\\ professor, etc...\end{tabular}       & \begin{tabular}[c]{@{}l@{}}film,movie\\ comedy,actor\\ series,tv,\\ directed,play...\end{tabular}       \\
\bottomrule
\end{tabular}
}
\\
\vspace{1mm}
\scalebox{0.73}{
\begin{tabular}{llllll}
          & \multicolumn{5}{c}{FewEvent}                                                                                                                                                                                                                                                                                                                                                                                                                                                                                                                                      \\
\midrule
cluster   & 1 (lowest entropy)                                                                                         & 2                                                                                                   & 3                                                                                                             & 33                                                                                                           & 34 (highest entropy)                                                                                            \\
\midrule
compoents & Military Service:150                                                                                       & \begin{tabular}[c]{@{}l@{}}Marry:162\\ Leadership:2\\ Place Lived:1\end{tabular}                    & \begin{tabular}[c]{@{}l@{}}Olympic Medal Honor:189\\ Education:3\\ Olympic Athlete Affiliation:3\end{tabular} & \begin{tabular}[c]{@{}l@{}}Leadership:19\\ Employment Tenure:9\\ Education:8\\ Place Lived:6\end{tabular}    & \begin{tabular}[c]{@{}l@{}}Sentence:14\\ Transfer Money:13\\ Charge Indict:13\\ Transport person:9\end{tabular} \\
\midrule
top words & \begin{tabular}[c]{@{}l@{}}soldier,medal,\\ war,honor,\\ received,sailor,\\ vietnam,killed...\end{tabular} & \begin{tabular}[c]{@{}l@{}}wife,birth,\\ died,maria,\\ child,king,\\ marriage,queen...\end{tabular} & \begin{tabular}[c]{@{}l@{}}olympics,medal,\\ gold,summer,\\ winner,relay,\\ competition,race...\end{tabular}  & \begin{tabular}[c]{@{}l@{}}event,speech,\\ triggered,words,\\ talk,rallies,\\ raise,appended...\end{tabular} & \begin{tabular}[c]{@{}l@{}}sentence,trigger,\\ charged,penalty,\\ crime,event,\\ prison,guilty...\end{tabular} \\
\bottomrule
\end{tabular}
}
\caption{Cluster explanations on other datasets. $^*$ Beowulf is the protagonist of an Old English epic poem of the same name, which is one of the most important works of Anglo-Saxon literature. }
\label{tab:explanation_app}
\end{table*}

%% file: tables/prompt_robustness_more.tex
\begin{figure*}[t]
    \centering
    \begin{subfigure}[b]{0.32\textwidth}
        \includegraphics[width=\textwidth]{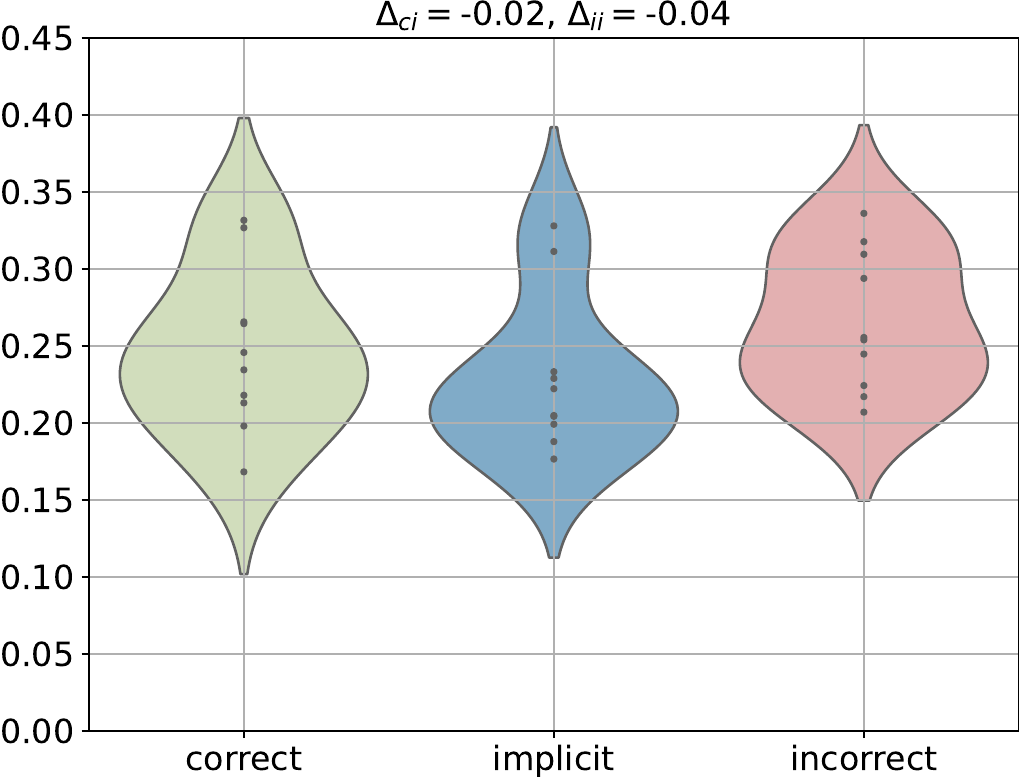}
    \end{subfigure}
    \begin{subfigure}[b]{0.32\textwidth}
        \subcaption*{Feedbacks}
        \includegraphics[width=\textwidth]{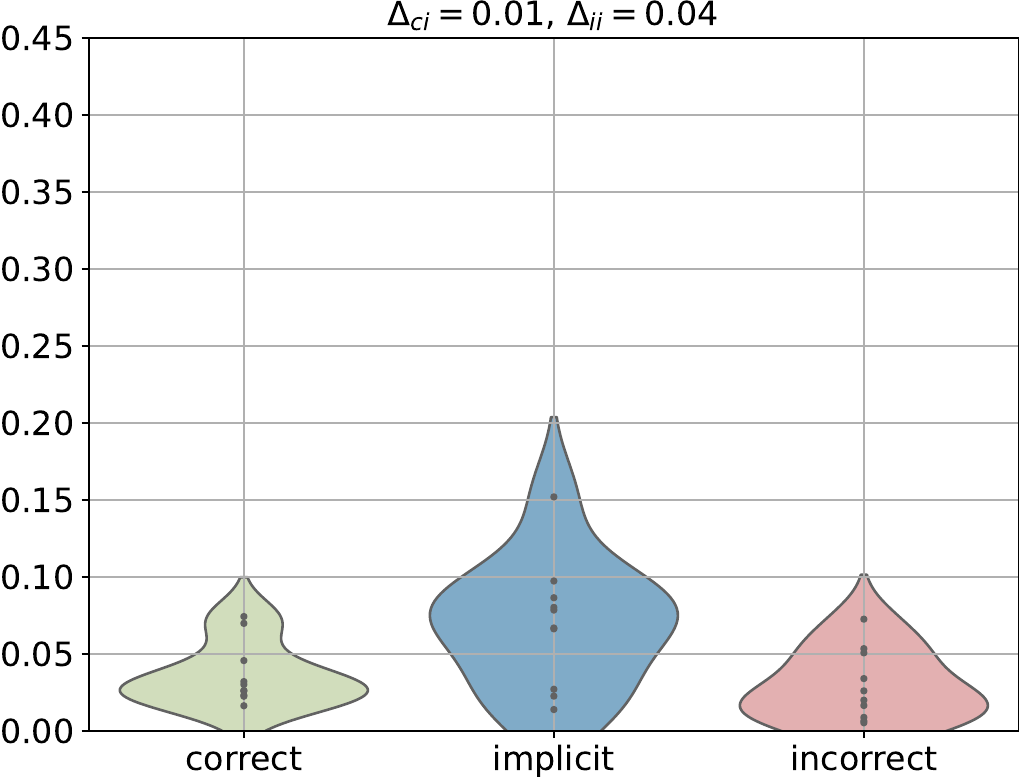}
    \end{subfigure}
    \begin{subfigure}[b]{0.32\textwidth}
        \includegraphics[width=\textwidth]{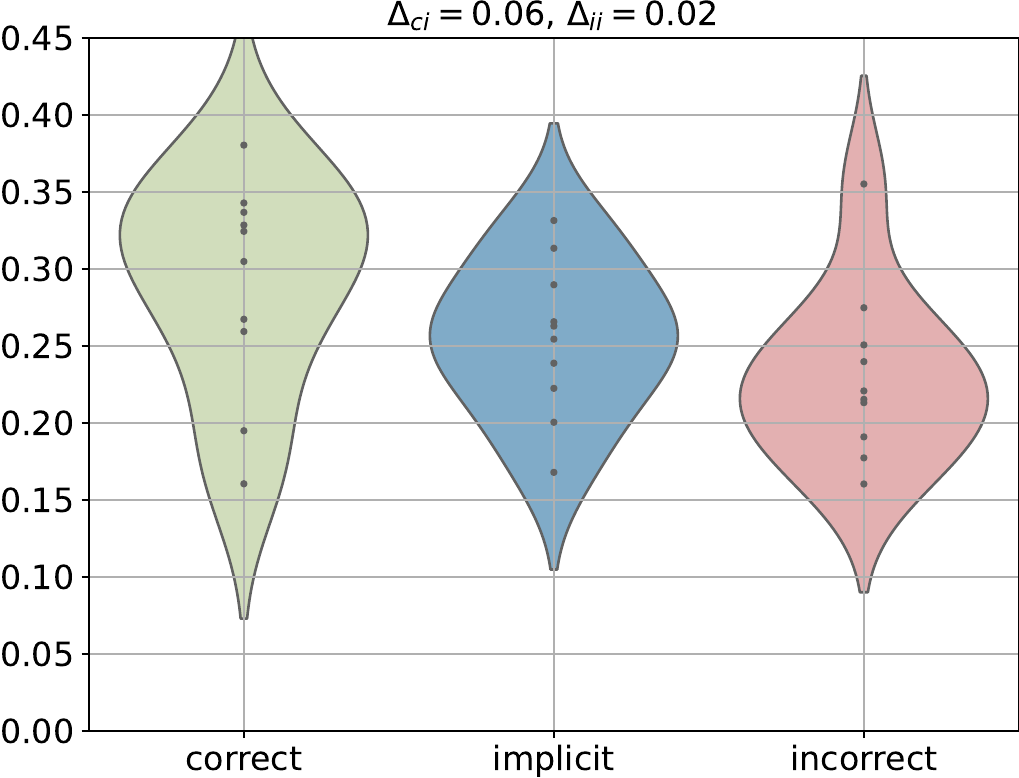}
    \end{subfigure}
    \\
    \vspace{1mm}
    \begin{subfigure}[b]{0.32\textwidth}
        \includegraphics[width=\textwidth]{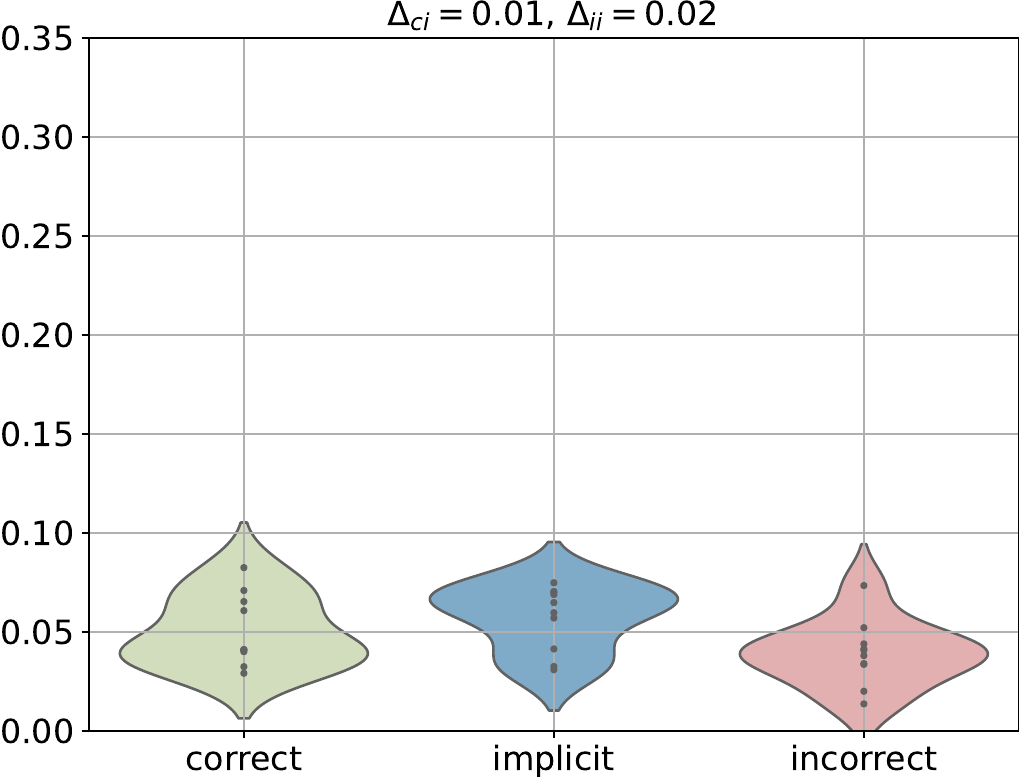}
    \end{subfigure}
    \begin{subfigure}[b]{0.32\textwidth}
        \subcaption*{RateMyProf}
        \includegraphics[width=\textwidth]{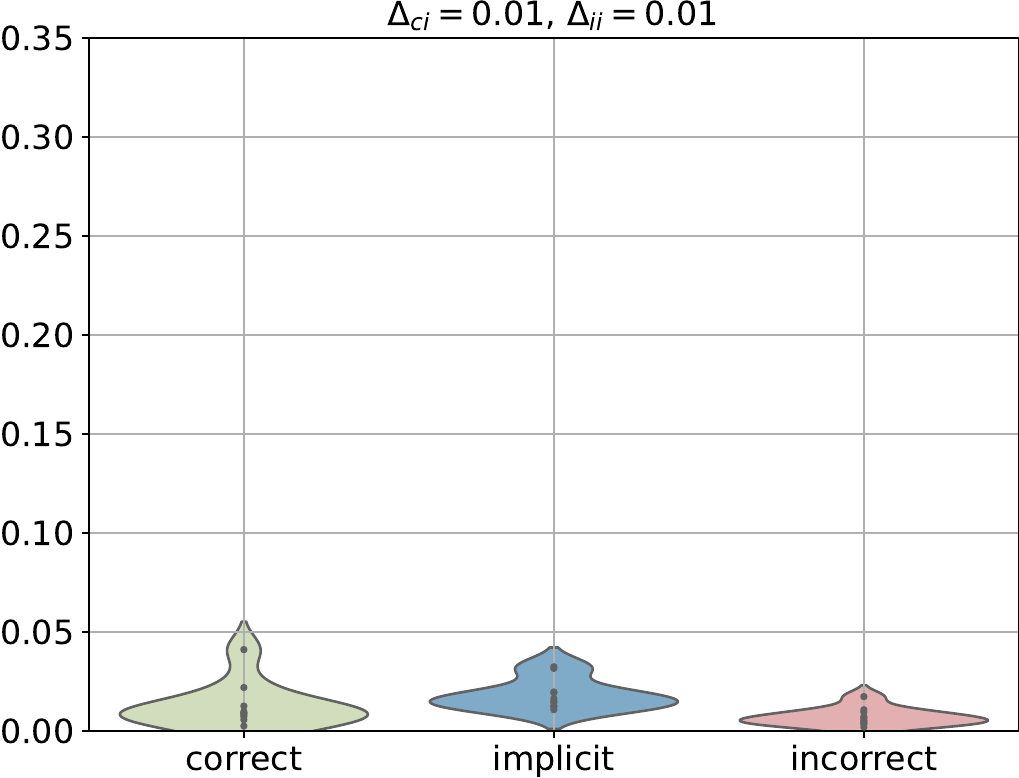}
    \end{subfigure}
    \begin{subfigure}[b]{0.32\textwidth}
        \includegraphics[width=\textwidth]{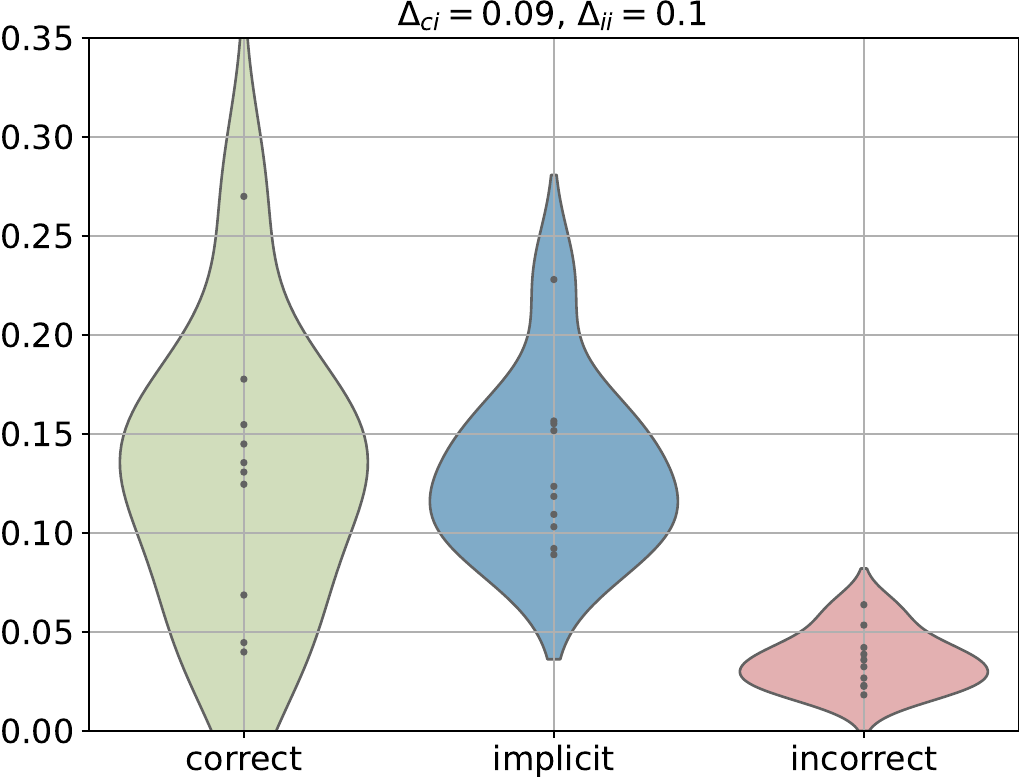}
    \end{subfigure}
    \\
    \vspace{1mm}
    \begin{subfigure}[b]{0.32\textwidth}
        \includegraphics[width=\textwidth]{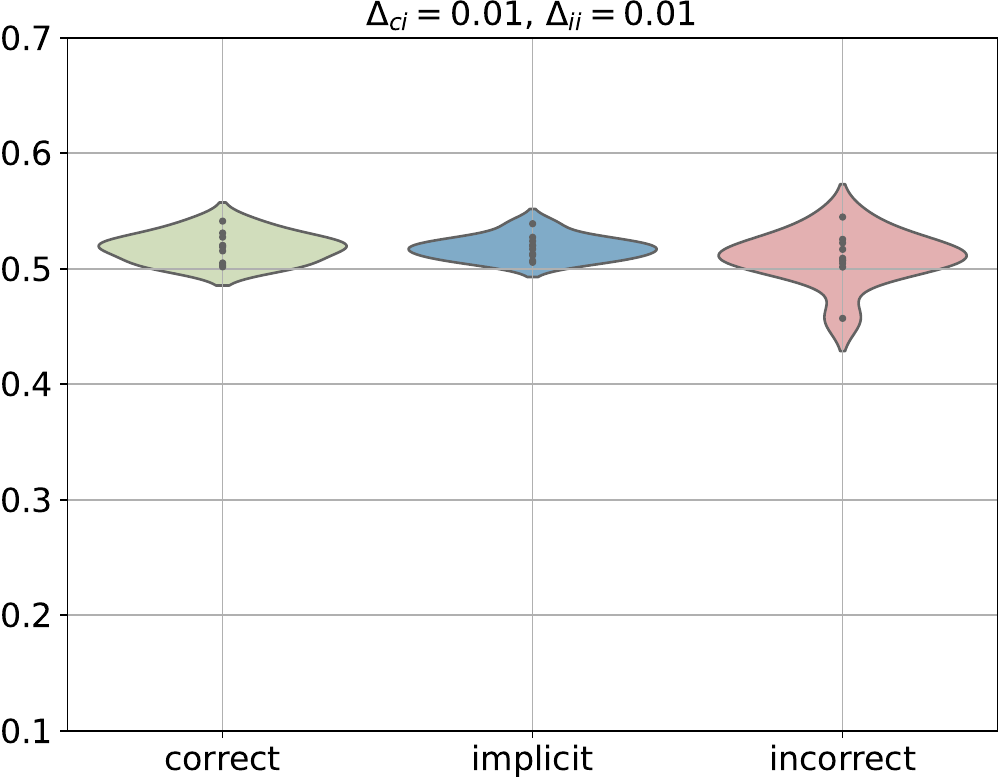}
    \end{subfigure}
    \begin{subfigure}[b]{0.32\textwidth}
        \subcaption*{FewRel}
        \includegraphics[width=\textwidth]{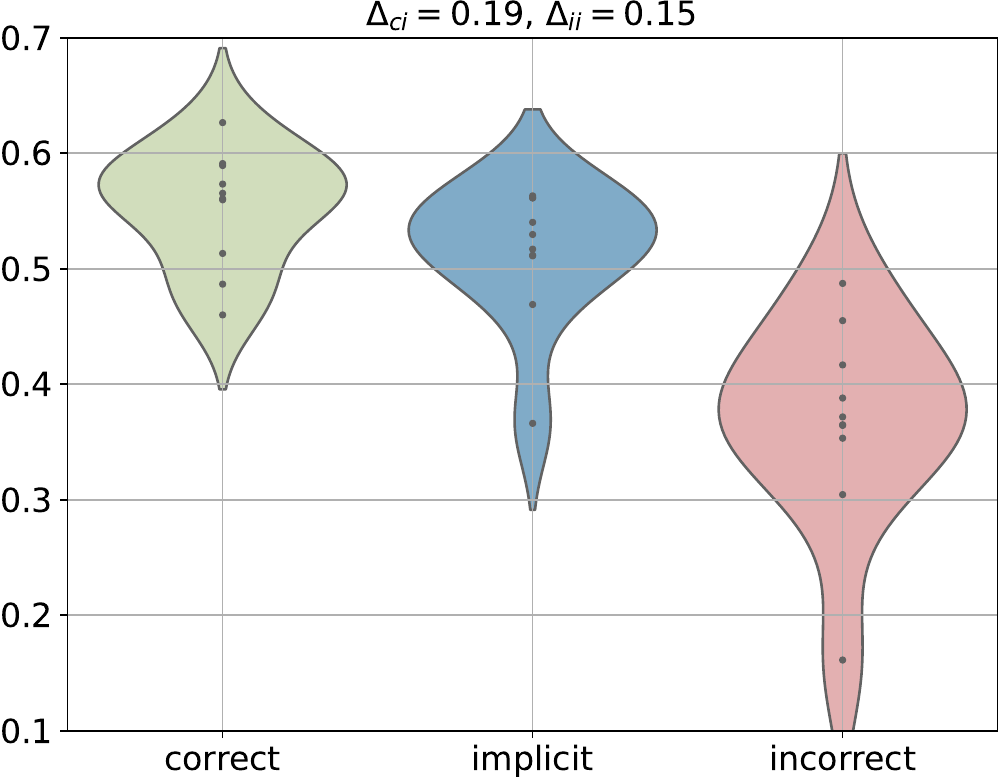}
    \end{subfigure}
    \begin{subfigure}[b]{0.32\textwidth}
        \includegraphics[width=\textwidth]{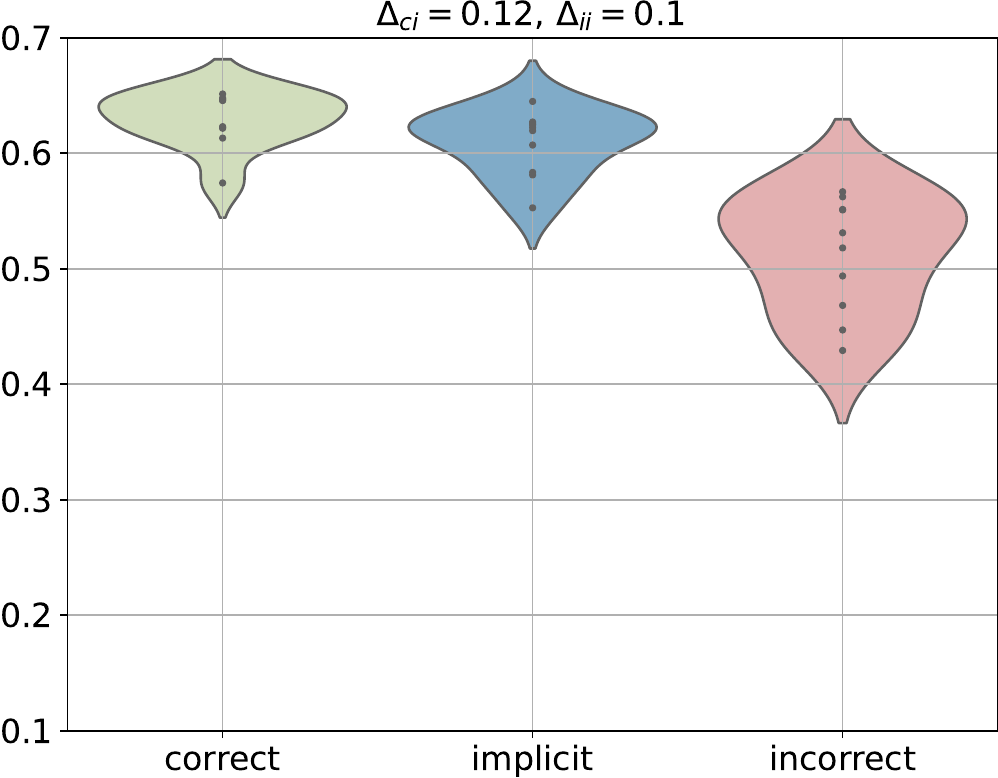}
    \end{subfigure}
    \\
    \vspace{1mm}
    \begin{subfigure}[b]{0.32\textwidth}
        \includegraphics[width=\textwidth]{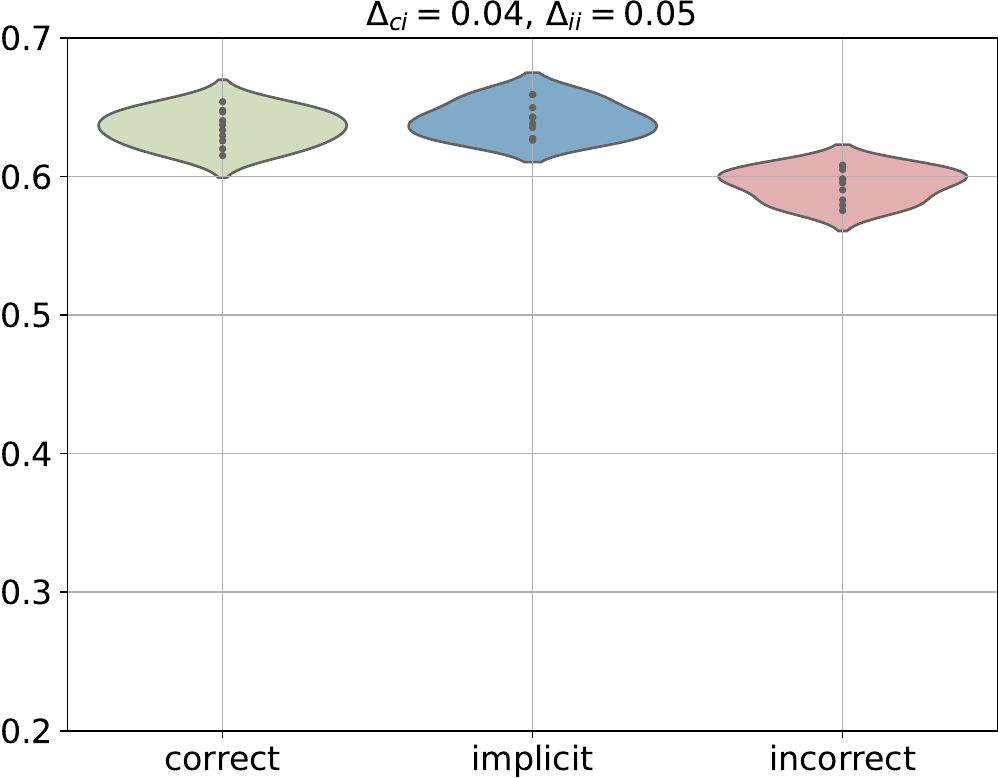}
        \caption*{\vspace{1mm}\\(a) instructor-large}
    \end{subfigure}
    \begin{subfigure}[b]{0.32\textwidth}
        \subcaption*{FewEvent}
        \includegraphics[width=\textwidth]{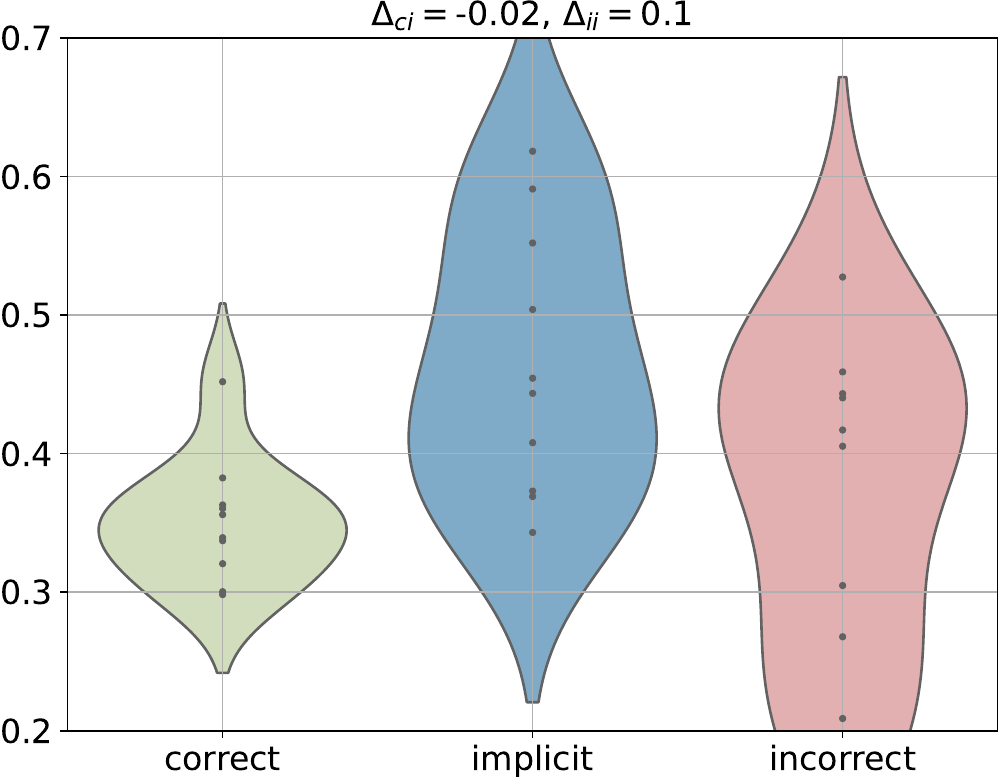}
        \caption*{\vspace{1mm}\\(b) llama-2-7b-chat (re-enc)}
    \end{subfigure}
    \begin{subfigure}[b]{0.32\textwidth}
        \includegraphics[width=\textwidth]{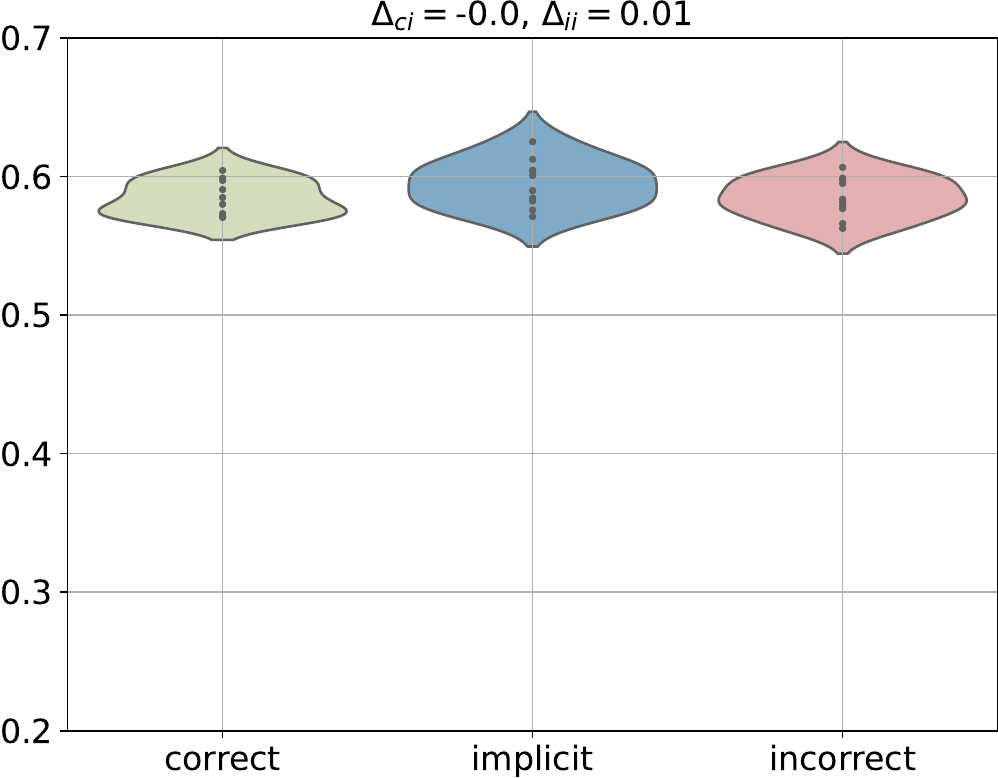}
        \caption*{\vspace{1mm}\\(c) llama-2-7b-\model (fst-gen)}
    \end{subfigure}
    \caption{Instruction robustness more results.}
    \label{fig:prompt_robustness_more}
\end{figure*}